\pdfoutput=1

\documentclass[11pt]{article}

\usepackage[final]{coling}

\usepackage{times}
\usepackage{latexsym}

\usepackage[T1]{fontenc}

\usepackage[utf8]{inputenc}

\usepackage{microtype}

\usepackage{inconsolata}

\usepackage{graphicx}

\usepackage{amsmath,amssymb}
\usepackage{multirow}
\usepackage{enumitem}

\usepackage{CJKutf8} 
\usepackage{CJK}

\usepackage{hyperref}
\usepackage[english]{babel}
\addto\captionsenglish{%
}
\addto\extrasenglish{%
}

\usepackage{subcaption}
\usepackage[whole]{bxcjkjatype}
\usepackage{comment}
\usepackage{enumitem}

\usepackage{color}

\title{Analyzing Continuous Semantic Shifts\\ with Diachronic Word Similarity Matrices
}

\author{
\normalsize {Hajime Kiyama}\textsuperscript{\rm 1} ~
{Taichi Aida}\textsuperscript{\rm 1} ~
{Mamoru Komachi}\textsuperscript{\rm 2} \\
\normalsize \textbf{{Toshinobu Ogiso}}\textsuperscript{3} ~
 \textbf{Hiroya Takamura}\textsuperscript{4} \and
 \textbf{Daichi Mochihashi}\textsuperscript{3,5}
\\[0.5em]
\normalsize \textsuperscript{1}Tokyo Metropolitan University
\textsuperscript{2}Hitotsubashi University \\
\normalsize \textsuperscript{3}National Institute for 
  Japanese Language and Linguistics ~
\normalsize \textsuperscript{4}National Institute of \\
\normalsize Advanced Industrial Science and Technology ~
\normalsize \textsuperscript{5}The Institute of Statistical Mathematics \\
\normalsize \texttt{\{kiyama-hajime,aida-taichi\}@ed.tmu.ac.jp}~
\texttt{mamoru.komachi@r.hit-u.ac.jp}\\
\normalsize \texttt{togiso@ninjal.ac.jp}~
\texttt{takamura.hiroya@aist.go.jp}~
\texttt{daichi@ism.ac.jp}
}

\begin{document}

\maketitle

\begin{abstract}
The meanings and relationships of words shift over time. 
This phenomenon is referred to as semantic shift.
Research focused on understanding \textit{how} semantic shifts occur over multiple time periods is essential for gaining a detailed understanding of semantic shifts.
However, detecting change points only between adjacent time periods is insufficient for analyzing detailed semantic shifts, and using BERT-based methods to examine word sense proportions incurs a high computational cost.
To address those issues, we propose a simple yet intuitive framework for \textit{how} semantic shifts occur over multiple time periods by leveraging a similarity matrix between the embeddings of the same word through time.
We compute a diachronic word similarity matrix using fast and lightweight word embeddings across arbitrary time periods, making it deeper to analyze continuous semantic shifts.
Additionally, by clustering the similarity matrices for different words, we can categorize words that exhibit similar behavior of semantic shift in an unsupervised manner.%
\footnote{The source code is available at \url{https://github.com/kiyama-hajime/acss-simmat}}
\end{abstract}

\section{Introduction}



\begin{figure*}[t]
\centering
\small
\begin{tabular}{ccc}
{\bf
\begin{tabular}[c]{@{}c@{}}
Input: The word embeddings \\ aligned for each period
\end{tabular}}
& 
{\bf
\begin{tabular}[c]{@{}c@{}}
1. Calculate the word similarity matrices \\ 
using word embeddings for each period.
\end{tabular}}
& 
{\bf
\begin{tabular}[c]{@{}c@{}}
2. Cluster the similarity \\
matrices of all words.
\end{tabular}} 
\\[-0.5em]
\begin{tabular}[c]{@{}c@{}}\\Dataset : Time 1-5\\ \\ \scalebox{0.11}{\includegraphics{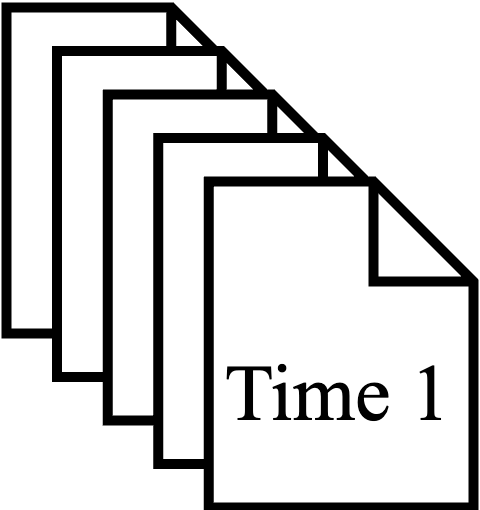}}\\ \\ 
\quad\qquad\Big\Downarrow~ {\it Learning} \\ \\ Word embeddings for each periods\\ \scalebox{0.11}{\includegraphics{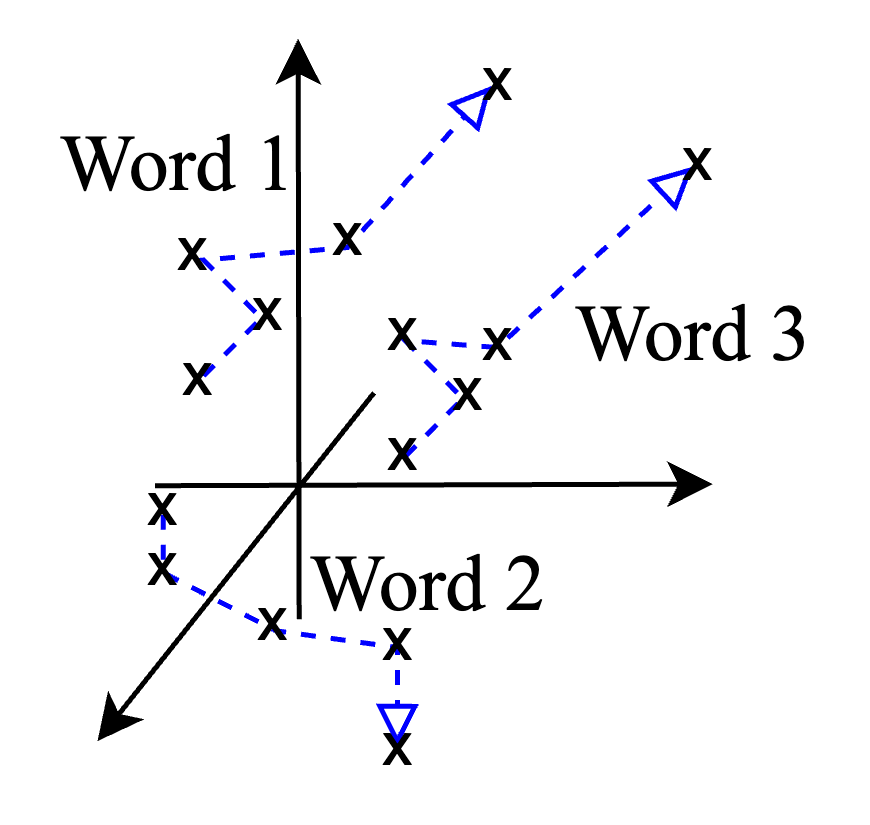}}\end{tabular} & \begin{tabular}[c]{@{}c@{}}\\ \scalebox{0.11}{\includegraphics{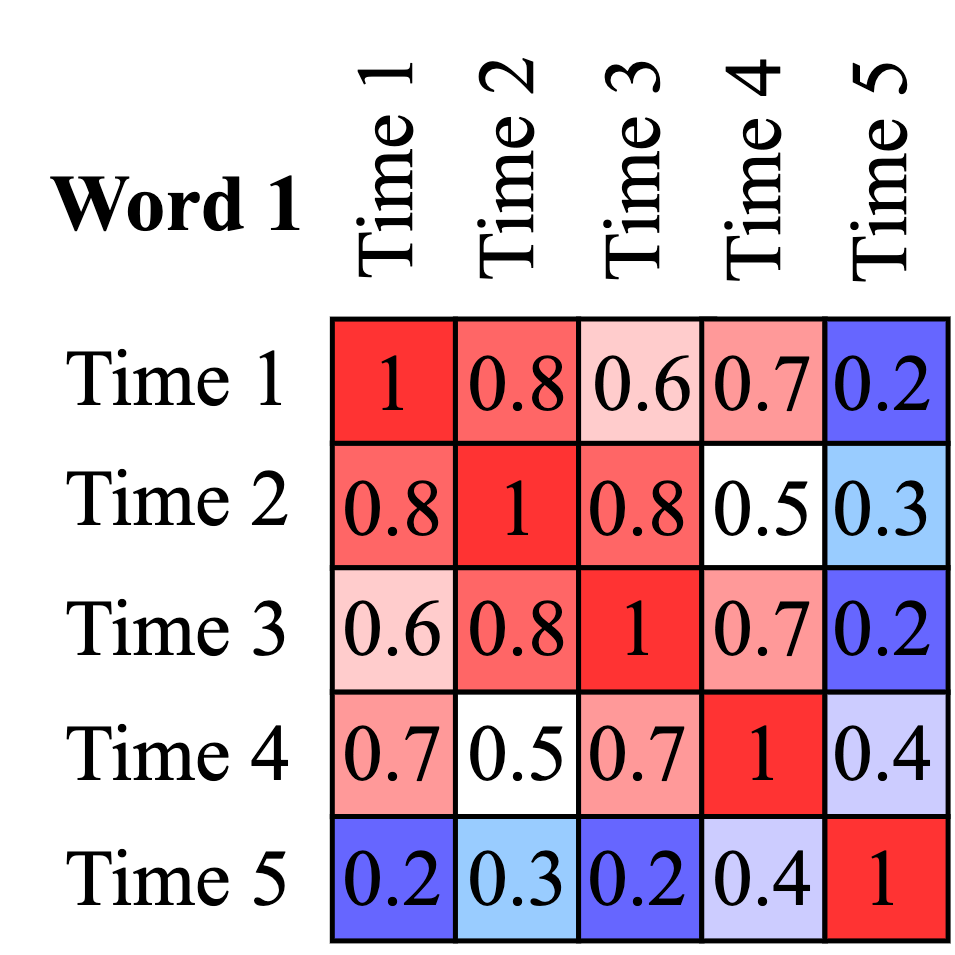}}\\ \scalebox{0.11}{\includegraphics{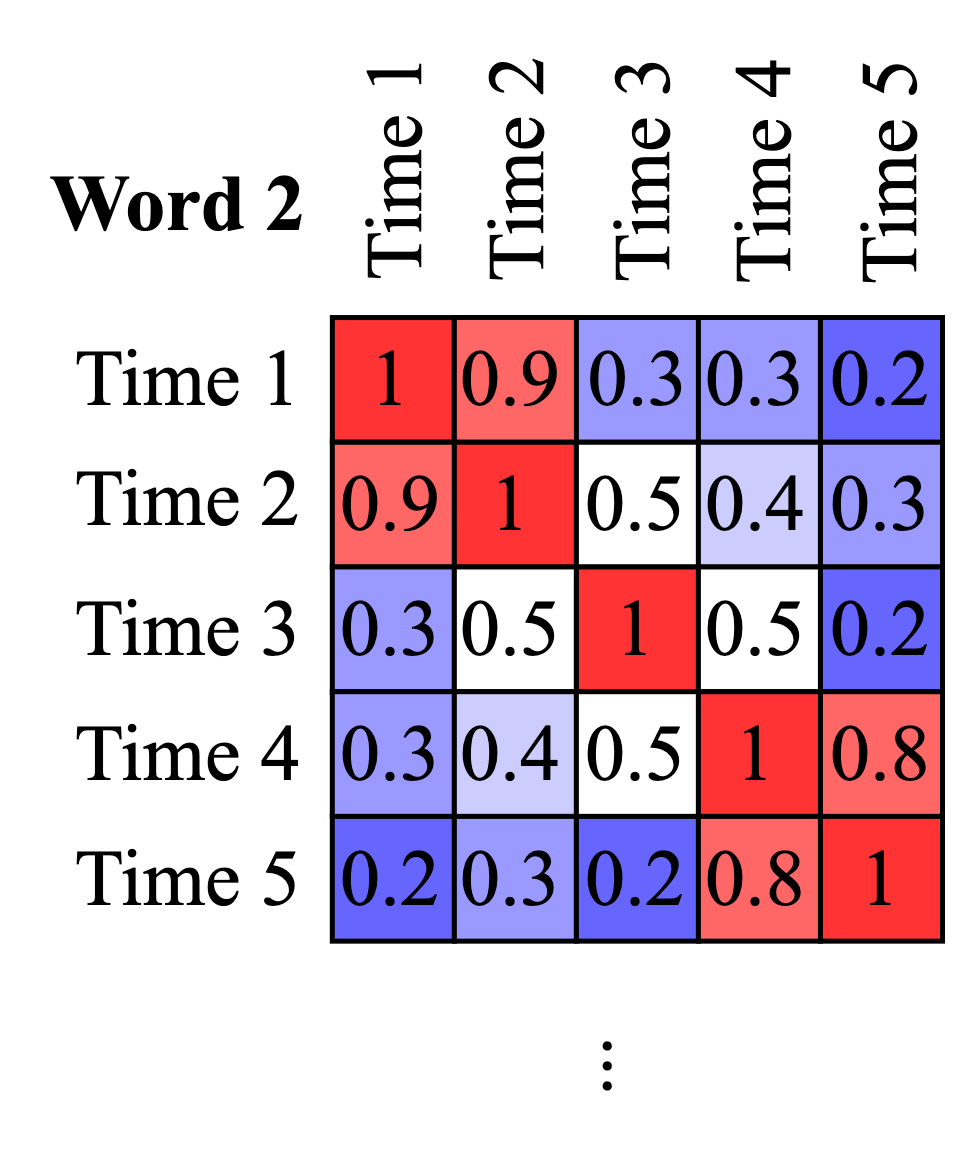}}\end{tabular} & \begin{tabular}[c]{@{}c@{}}\\Input : Similarity matrices of all words\\ \\ \scalebox{0.11}{\includegraphics{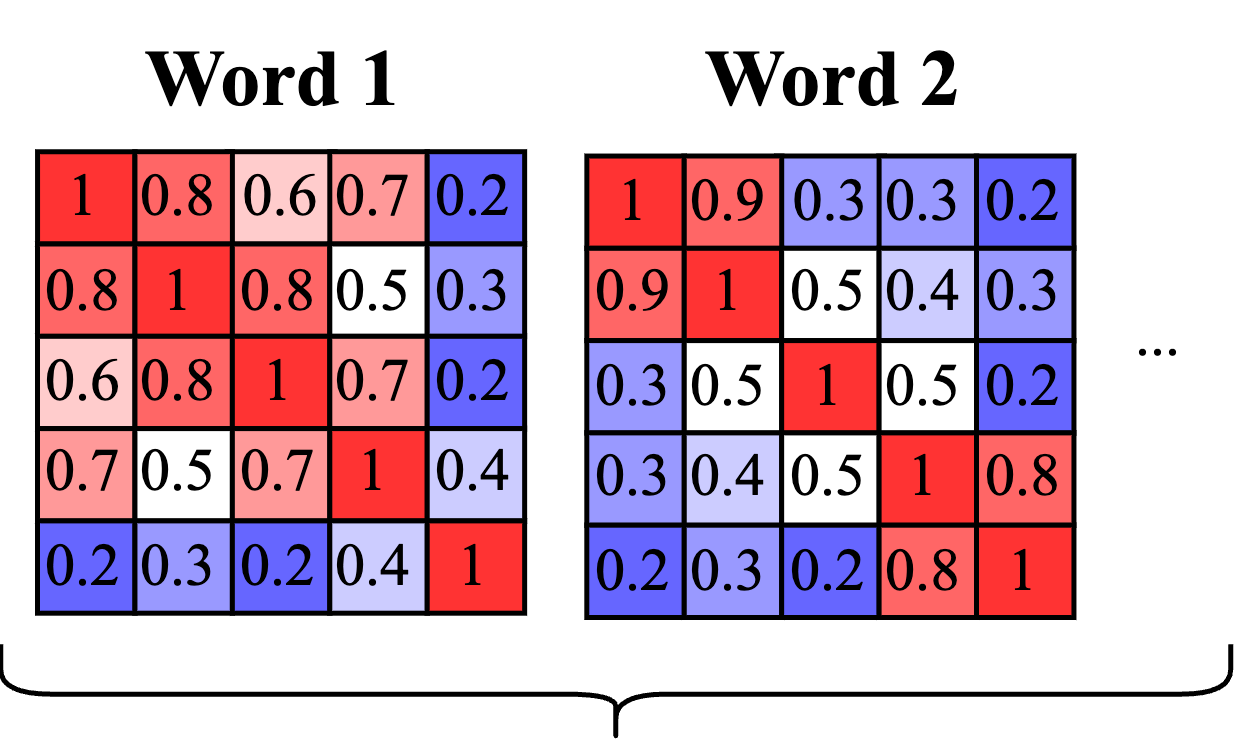}}\\ \\ \,\,\,\,\,\,\,\,\,\,\,\,\,\,\,\,\,\,\,\,\,\,\,\,\,\,\Big\Downarrow~ {\it Clustering} \\ \scalebox{0.1}{\includegraphics{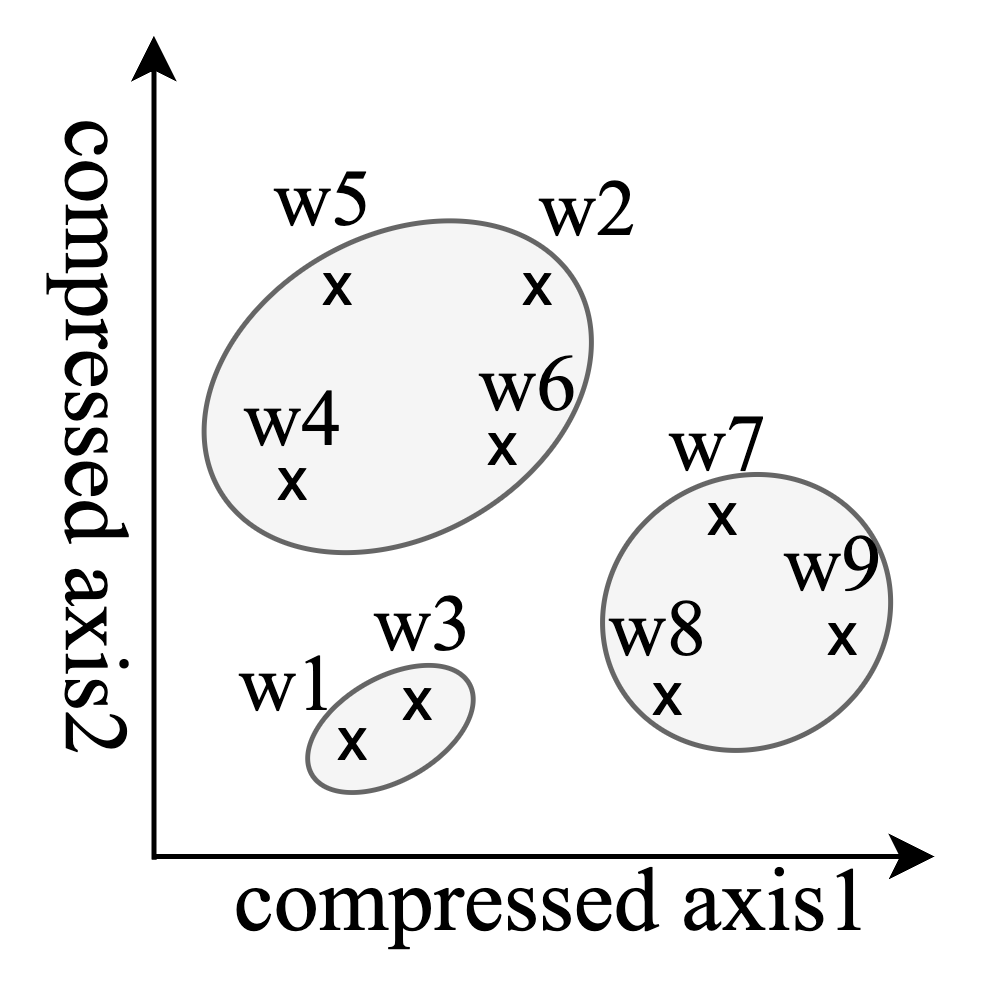}}\end{tabular}
\end{tabular}
\vspace{-1.3em}
\caption{A framework for analysing diachronic semantic shifts using similarity matrices:
1. Calculate the word similarity matrix for a target word using word embeddings trained for each period.
2. Perform analyses such as clustering on the  similarity matrices for all words.}
\label{fig:intro}
\end{figure*}

The word embedding captures the meaning of words based on the distributional hypothesis~\cite{distributional-harris,distributional-firth}, so changes in context are reflected as shifts of the word embedding.
The phenomenon where the meanings and relationships of words shift over time is referred to as \textit{semantic shift} \cite{kutuzov-survey-2018, Periti-survey-2024,desa-2024-surveycharacterizationsemanticchange}. 
For example, the word \textit{horn} initially meant ``animal horn'', but over time, it has acquired the meaning of ``brass instrument''~\cite{stern-semantic-change}. 
Many studies in natural language processing have computationally detected such semantic shifts.

There are two major questions in the research of semantic shift: \textit{what words have their meaning shifted between two periods} and \textit{how the meanings of words have shifted over multiple time periods}~\cite{periti-tahmasebi-2024-towards}.
For the first question, measuring the degree of semantic shift between two time periods are commonly used~\cite{laicher-etal-2021-explaining,10.1145/3488560.3498529,rosin-radinsky-2022-temporal,cassotti-etal-2023-xl, periti-tahmasebi-2024-systematic,aida2024semanticdistancemetriclearning, periti2024analyzingsemanticchangelexical}. Although there are manually annotated datasets~\cite{schlechtweg-etal-2020-semeval}, it remains challenging to analyze how these semantic shifts have occurred.
For the second question, there are research of change point detection~\cite{change-detection-2014} and analysis of proportions of word sense in BERT-based approach~\cite{hu-etal-2019-diachronic,giulianelli-etal-2020-analysing}. 
However, even though change points between adjacent time periods can be identified, it remains unclear whether the sense reverts to the original or transitions to a new one.
While it is possible to cluster multiple word embeddings using BERT-based embeddings, computational limitations restrict the number of target words.

In contrast, we propose a simple yet intuitive framework to address the second question by diachronic word similarity matrices. 
\autoref{fig:intro} provides an overview.
The framework involves the following steps:
As input, we prepare word embeddings aligned across different time periods.
\begin{itemize}[noitemsep,topsep=2pt]
 \item[(i)] Calculation of diachronic word similarity matrices. 
The diachronic word similarity matrices provide insights into the dynamics of semantic shifts for individual words.
 \item[(ii)] Clustering of the similarity matrices for all words.
The clustering results summarize the semantic shift dynamics across words in an unsupervised manner.
\end{itemize}

The contributions of this study are as follows:
(i) we analyzed semantic shifts across arbitrary time periods using diachronic word similarity matrices. 
By identifying high similarity regions within the similarity matrix and calculating the differences in Positive Pointwise Mutual Information (PPMI) for each relevant period, we enabled detailed analysis of semantic shifts.
(ii) clustering the word similarity matrices allowed us to group words with similar behavior such as high-similarity regions split into two distinct groups in an unsupervised setting. 
This study used fast and lightweight word embeddings to increase the number of target words for analysis. 
In addition, we conducted a shift schema classification task using pseudo data to quantitatively verify the validity of the proposed framework.

\section{Related Work}
We will discuss research on semantic shifts across multiple time periods in \autoref{subsec:related-multi} and research on applying a generalized similarity matrix, or Gram matrix, to time-series data in \autoref{subsec:related-sim}.

\subsection{Semantic Shift over Multiple Periods}
\label{subsec:related-multi}
Research on semantic shift analysis over multiple periods follows two approaches: one that assumes specific types of semantic shifts and one without such assumptions.

First, we show researches of one that reveal and classify specific types of semantic shifts.
\citet{hamilton-etal-2016-diachronic} explore the statistical laws of semantic shift, focusing on frequency and polysemy. 
\citet{feltgen-freq-pattern-2017} computationally reveal the S-curve frequency pattern in semantic shifts. 
\citet{shoemark-etal-2019-room} define seven shift schemas as part of a framework for evaluating semantic shift detection. 
\citet{cassotti2024usingsynchronicdefinitionssemantic} define a task for classifying these types and propose a method for categorizing types of semantic shifts. 
\citet{baes-etal-2024-multidimensional} have analyzed semantic shifts by defining the factors or dimensions that drive these shifts.

Next, we show researches of unsupervised types analyses of semantic shifts.
\citet{change-detection-2014} propose a method to detect the period during which a semantic shift occurred by calculating the distance between the embeddings of two adjacent time periods.
\citet{yao-evolving-2018} analyze trends in meaning driven by social factors by incrementally training word embeddings for each period. 
\citet{hu-etal-2019-diachronic} analyze the process of semantic shift from an ecological perspective. 
\citet{inoue-etal-2022-infinite} propose a Bayesian method using topic models to estimate the number of senses across multiple periods and track their shifts. 

The methods discussed in \autoref{subsec:related-multi} face certain limitations: (i) They assume certain types of semantic shifts in their analysis, (ii) They are analyzing semantic shifts between adjacent time periods, and (iii) They limit the target words and calculate word senses for analysis. 
To address those issues, we propose a semantic shift analysis framework using diachronic word similarity matrices.
This framework does not assume any specific type of semantic shift and allows for analysis across arbitrary time periods.
By using fast and lightweight word embeddings, it enables the selection of a large number of target words.

\subsection{Gram Matrix for Time-Series Data}
\label{subsec:related-sim}
Calculating similarity matrices is equivalent to computing a Gram matrix where similarity is treated as an inner product in kernel methods. \footnote{While it is possible to calculate similarity using various kernel methods, this study investigates only simple similarity measures such as cosine similarity and Euclidean distance, motivated by the need for lightweight computation.}
There are numerous studies that transform time-series data into Gram matrices for classification purposes. These studies use various data sources such as 3D human movements~\cite{zhang-gram-3d,kacem-gram-matrix-3d}, eye movements~\cite{qiu-gram-nystagmography}, and sound events~\cite{antonio-gram-sound-2021}. The resulting Gram matrices are often used as features for classification with supervised methods like SVM or CNN.

In natural language processing, Gram matrices have been utilized to investigate the tendencies of language models fine-tuned for specific time periods. 
\citet{nylund-etal-2024-time} have proposed time vectors, where model weight vectors corresponding to different periods are used to track how model features change over time.
In this research, we calculate diachronic word similarity matrices using word embeddings from different time periods and cluster the similarity matrices.
Instead of using time vectors, word vectors from each period are utilized. 

\section{Diachronic Word Similarity Matrices for Semantic Shift}
We explain a simple yet intuitive framework for semantic shifts analysis by similarity matrices (\autoref{fig:intro}).
First, we calculate diachronic word similarity matrices using the word embeddings from each time period (\autoref{subsec:frame-sim}).
Then, we perform clustering based on the similarity matrices calculated for all words (\autoref{subsec:frame-cluster}).
From the clustering results, we can group words that exhibit similar behaviors in their similarity matrices.

\begin{figure*}[tb]
    \begin{tabular}{c}
    \centering
    \begin{minipage}{0.5\hsize}
        \centering
        \includegraphics[keepaspectratio, scale=0.05]{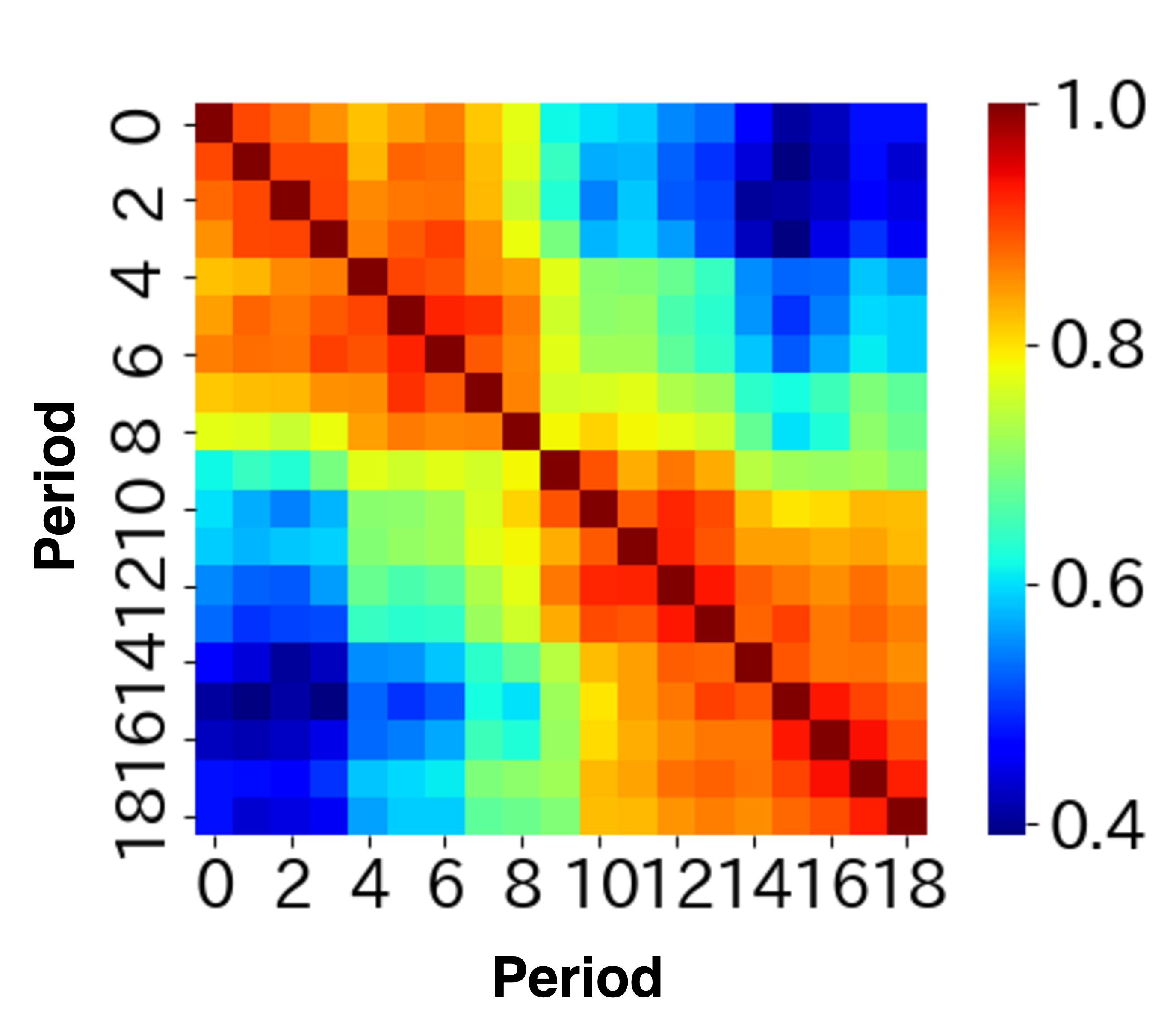}
        \vspace{-1ex}
        \subcaption{Cosine similarity matrix of ``record'' vectors in COHA.}
        \label{fig:cos}
    \end{minipage}
    \begin{minipage}{0.5\hsize}
        \centering
        \includegraphics[keepaspectratio, scale=0.05]{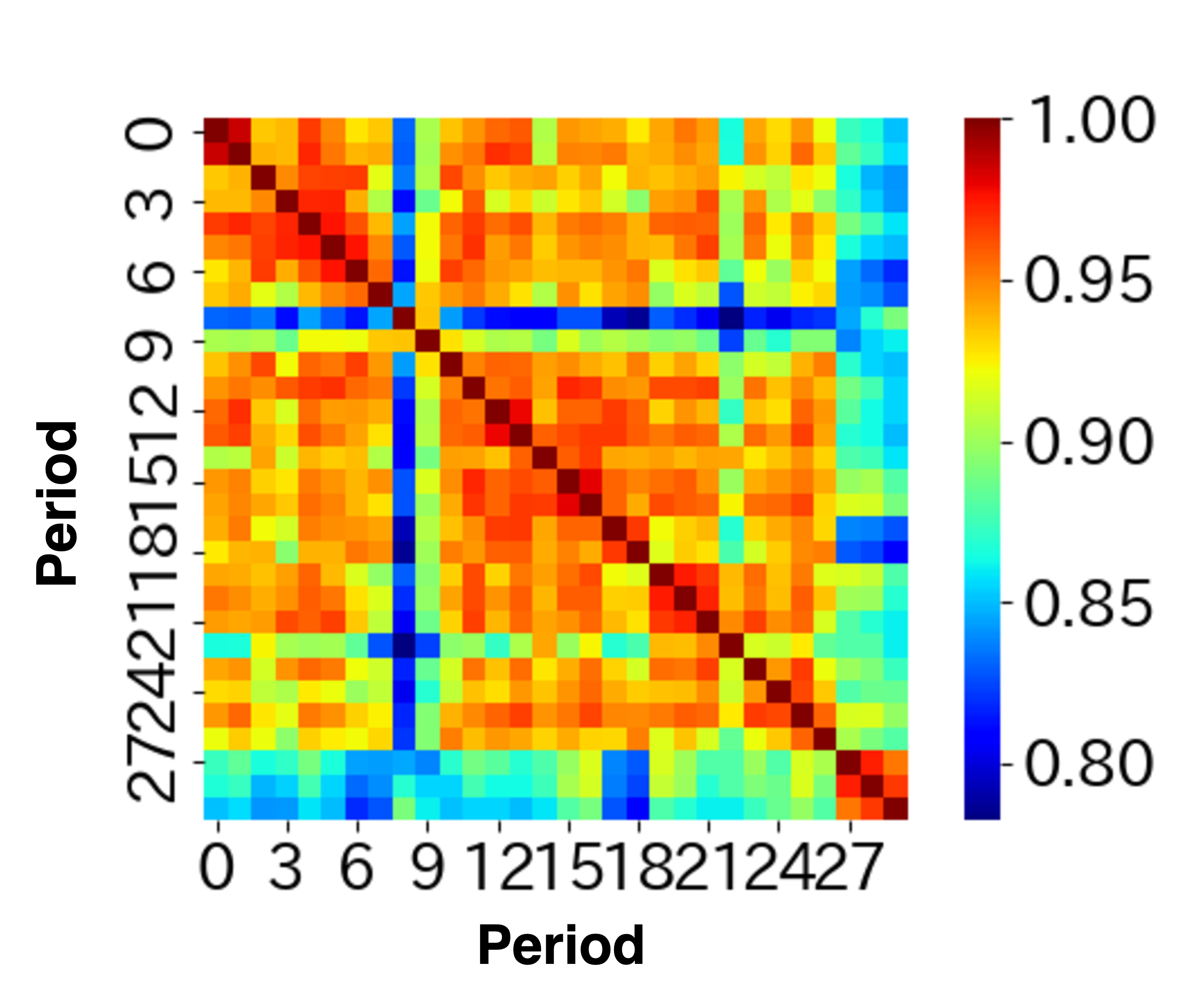}
        \vspace{-1ex}
        \subcaption{Cosine similarity matrix of ``president'' vectors in COCA.}
        \label{fig:cos_coca}
    \end{minipage}
    \end{tabular}
    \caption{The visualization of the diachronic word cosine similarity matrix for the word ``record'' and ``president'' by PPMI-SVD joint. 
    It is evident that clusters and spikes across time, indicating two types of semantic shifts (linguistic and social), have been successfully represented.
    }
    \label{fig:cos_all}
\end{figure*}

\paragraph{Input: Word Embeddings}
The input to the framework consists of word embeddings from each period. 
These embeddings must either be obtained from the same model to ensure consistency or have their dimensions aligned across different periods to facilitate comparison. 
The constraint on input exists because we aim to cluster words that exhibit similar trajectories by mapping the temporal transitions of word embeddings onto the same space.
In this study, we use PPMI-SVD joint~\cite{aida-etal-2021-comprehensive} to efficiently compute and prepare a large number of target words.
Positive Pointwise Mutual Information (PPMI) is an indicator that measures the degree of association between two words.
Here, let the target word be $w$, the context word be $c$, and their (co-occurrence) probabilities be $p(w)$, $p(c)$, and $p(w,c)$ respectively. 
The PPMI matrix for period $t$, $M^t \in \mathbb{R}^{W \times C}$, is defined as 
\begin{align*}
M^{(t)}_{wc} = \max \left( \log \dfrac{p(w,c)}{p(w)p(c)},0 \right).
\end{align*}
The PPMI matrices obtained here, when compressed through singular value decomposition (SVD) as $M^{(t)}=U\Sigma V^T$, become equivalent to word embeddings of skip-gram with negative sampling (SGNS)~\cite{levy-2014-matrix}\footnote{Following \citet{aida-etal-2021-comprehensive}, we also employ the original PPMI. 
Although SGNS has been shown to correspond to a shifted form of PMI, \citet{levy-etal-2015-improving} demonstrated that incorporating this shift does not improve performance when applied to PMI matrix factorization. } as $W = U\Sigma^{1/2}$ where $U$ and $V$ are orthogonal matrices, and $\Sigma$ is a diagonal matrix consisting of singular values of $M$. 
PPMI-SVD joint allows for obtaining embeddings with aligned dimensions across all time periods by performing the SVD-based compression simultaneously for all periods.

\subsection{Calculation of Diachronic Word Similarity Matrices}
\label{subsec:frame-sim}
To compute similarities across multiple time periods (\autoref{fig:intro}(i)), we first obtain $D$-dimensional word embeddings $e_t(w) \in \mathbb{R}^D$ for a given word $w$ at each period $t \in \{1, 2, ..., T\}$.
Let the function $\mathrm{sim}(\cdot,\cdot)$ return a similarity score. 
The similarity matrix $S(w) \in \mathbb{R}^{T \times T}$ for word $w$ is defined as
\begin{align*}
    S_{ij}(w) = \mathrm{sim}(e_{i}(w),e_{j}(w)).
\end{align*}
In this study, we adopt cosine similarity. 
One advantage of using similarity matrices across arbitrary time periods is that it facilitates the analysis of how semantic shifts have occurred. 
This method makes semantic shift easier to interpret compared to traditional change point detection methods that focus only on adjacent periods (\autoref{subsec:exp-pseudo_embed_v_simmat}).

\subsection{Clustering of Diachronic Word Similarity Matrices}
\label{subsec:frame-cluster}
When clustering the obtained similarity matrix (\autoref{fig:intro}(ii)), the upper triangular part of the similarity matrix is extracted and normalized to obtain a serialized vector. 
This preprocessing step is taken because the similarity matrix is symmetric and our focus is on analyzing the shifts in the similarity matrix. 
Analyzing the similarity matrices within each cluster allows for a clearer understanding of the semantic shifts within the cluster, thereby enhancing the interpretability of the clustering outcome.
In this study, we use hierarchical clustering.\footnote{To determine the method with the highest classification performance, similarity methods, elements of similarity matrices, clustering methods and the presence or absence of standardization are investigated in \autoref{subsec:pseudo-class}.}

\section{Experiment: Real Data}
\label{chap:simmat}
We demonstrate that diachronic word similarity matrices are beneficial for analyzing semantic shifts. 
First, we describe experimental setup of English corpora in \autoref{subsec:simmat-set}.
Next, we visualize the actual similarity matrices, demonstrating that it allows deeper analysis even across different time slices than using only adjacent time periods in \autoref{subsec:exp-pseudo_embed_v_simmat}.
Additionally, we show that the similarity matrix itself can be analyzed using PPMI in \autoref{subsec:exp-pseudo_embed_v_simmat}.
Finally, we present the clustering results for the similarity matrices in English corpora in \autoref{subsec:real-class}.
\footnote{The analysis was also conducted in Japanese, and the results are presented in \autoref{sec:appendix:japan}. It was found that, similar to the English experiments, the analysis using similarity matrices is also beneficial in the Japanese experiments. }

\begin{table*}[t]
\small
\centering
\begin{tabular}{llll}
$t_1$ (target year) & $t_2$   & co-occurring words   & sense interpretation           \\ \hline
\rule{0pt}{1.5em}%
1840 & 1940
&
\begin{tabular}[c]{@{}l@{}}miracle, inspiration, tradition, pen, chapter, \\contemporary, interpret, preserve, translate, journal\end{tabular} & record in memorizing things 
\\ \hline
\rule{0pt}{1.5em}%
1940 & 1840
& \begin{tabular}[c]{@{}l@{}}attendance, concert, sale, speed, employment,\\ consistent, error, moderate, arrest, tune\end{tabular}            & record in medium for playing sound      \\ \hline
\end{tabular}
\caption{Top 10 words and their meanings sorted by the differences in PPMI ($\mathcal{N}_{10}^{(t_2 \rightarrow t_1)}$) for the word ``record'' between 1840 and 1940 learned from COHA. }
\label{tab:record}
\end{table*}

\begin{table*}[t]
\small
\centering
\begin{tabular}{llll}
$t_1$ (target year) & $t_2$   & co-occurring words   & sense interpretation or event              \\ \hline
\rule{0pt}{1.5em}%
1991 & 2017
& 
\begin{tabular}[c]{@{}l@{}}dan, republic, marketing, resolution, initiative,\\ peace, proposal, coalition, founder, approve\end{tabular}            & president in normal uses \\ \hline
1998 & 1991
& \begin{tabular}[c]{@{}l@{}}fox, rice, video, conservative, democracy, \\texas, mexican, supreme, walker, andrew\end{tabular} & \begin{tabular}[c]{@{}l@{}}documentary of U.S. President\\ Andrew Jackson (event)  \end{tabular}  \\ \hline
\rule{0pt}{1.5em}%
2012 & 1991
& \begin{tabular}[c]{@{}l@{}}ohio, convention, debate, joe, criticism,\\ tax, immigration, fiscal, voter, lincoln\end{tabular} & presidential election in 2012 (event)      \\ \hline
\rule{0pt}{1.5em}%
2017 & 1991
& \begin{tabular}[c]{@{}l@{}}joe, investigation, border, moon, defend,\\ mike, korea, lawyer, counsel, investigate\end{tabular} & president in Trump administration (event)      \\ \hline
\end{tabular}
\caption{Top 10 words and their meanings sorted by the differences in PPMI ($\mathcal{N}_{10}^{(t_2 \rightarrow t_1)}$) for the word ``president'' across four time periods: 1991, 1998, 2012 and 2017 learned from the COCA.}
\label{tab:president}
\end{table*}

\subsection{Experimental Setup}
\label{subsec:simmat-set}
We use COHA~\cite{coha,alatrash-etal-2020-ccoha} and COCA~\cite{coca} as datasets.\footnote{Details of the experimental setup, including corpus statistics, are provided in  \autoref{sec:appendix:real-setting}.} 
COHA is an English historical corpus, segmented into 10-year periods from 1830 to 2010, resulting in subcorpora for 19 time periods. 
We did not use the data of 1820s because the data size was too small.
COCA is an English contemporary corpus, segmented into 1-year periods from 1990 to 2019, resulting in subcorpora for 30 time periods. 
For all datasets, the target words are those that appear more than 100 times in each period. 
The numbers of target words in COHA and COCA are 3,231 and 2,805, respectively.
We used a 100-dimensional PPMI-SVD joint for word embeddings and cosine similarity for measuring similarity.

In the clustering step, we use the combination of feature, preprocessing and clustering methods that produced the best results in the pseudo-data experiments (cosine similarity + upper triangular matrix  + hierarchical clustering + standardization) in \autoref{chap:pseudo}. 
We set the threshold for hierarchical clustering to 8 in COHA and 30 in COCA.\footnote{Setting the optimal threshold in clustering is challenging. In this study, the threshold for clustering was manually determined. 
This decision was made because the calculation of the silhouette score indicated that the optimal number of clusters was 2, after which the score tended to decrease monotonically. 
The figures showing the silhouette scores calculated for hierarchical clustering and K-means are provided in \autoref{sec:appendix:class}.}

\subsection{Analysis of Diachronic Word Similarity Matrices}
\label{subsec:exp-pseudo_embed_v_simmat}
We present qualitative results of similarity matrix visualizations (\autoref{subsec:vis-sim}) and quantitative analysis based on the differences in PPMI (\S \,\ref{subsec:simmat-ppmi}).

\subsubsection{Visualization of Diachronic Word Similarity Matrix}
\label{subsec:vis-sim}
We visualize the diachronic word similarity matrices for COHA and COCA. 
By visualizing the similarity matrix, we demonstrate that semantic shift analysis can be conducted even with different time slices, and that tracking semantic shifts across arbitrary time periods allows for more detailed insights than focusing only on adjacent periods.

\autoref{fig:cos} shows the similarity matrix of word embeddings for the word ``record'' learned from COHA.
The word ``record'' was selected because it is known to have undergone semantic shift.
The similarity matrix indicates that, around period 9 (1930), high-similarity regions split into two distinct groups, suggesting that the meaning of ``record'' shifted around 1930.

\autoref{fig:cos_coca} presents the similarity matrix for the word ``president''  learned from COCA. 
The word ``president'' was selected because it was hypothesized that its meaning might have shifted due to social factors, particularly in the context of U.S. presidential elections. Analysis of the similarity matrix reveals that the data generally splits into two regions: one spanning from period 0 (1991) to period 26 (2016), and the other from period 27 (2017) to period 29 (2019). 
Additionally, spikes are observed in period 8 (1998) and period 22 (2012), suggesting significant shifts likely related to the documentary of the seventh U.S. president Andrew Jackson (1998) and presidential elections (2012)\footnote{Barack Hussein Obama II vs Willard Mitt Romney}, which are reflected in the language of the texts.

\begin{figure*}[t]
    \centering
    \includegraphics[keepaspectratio, scale=0.12]{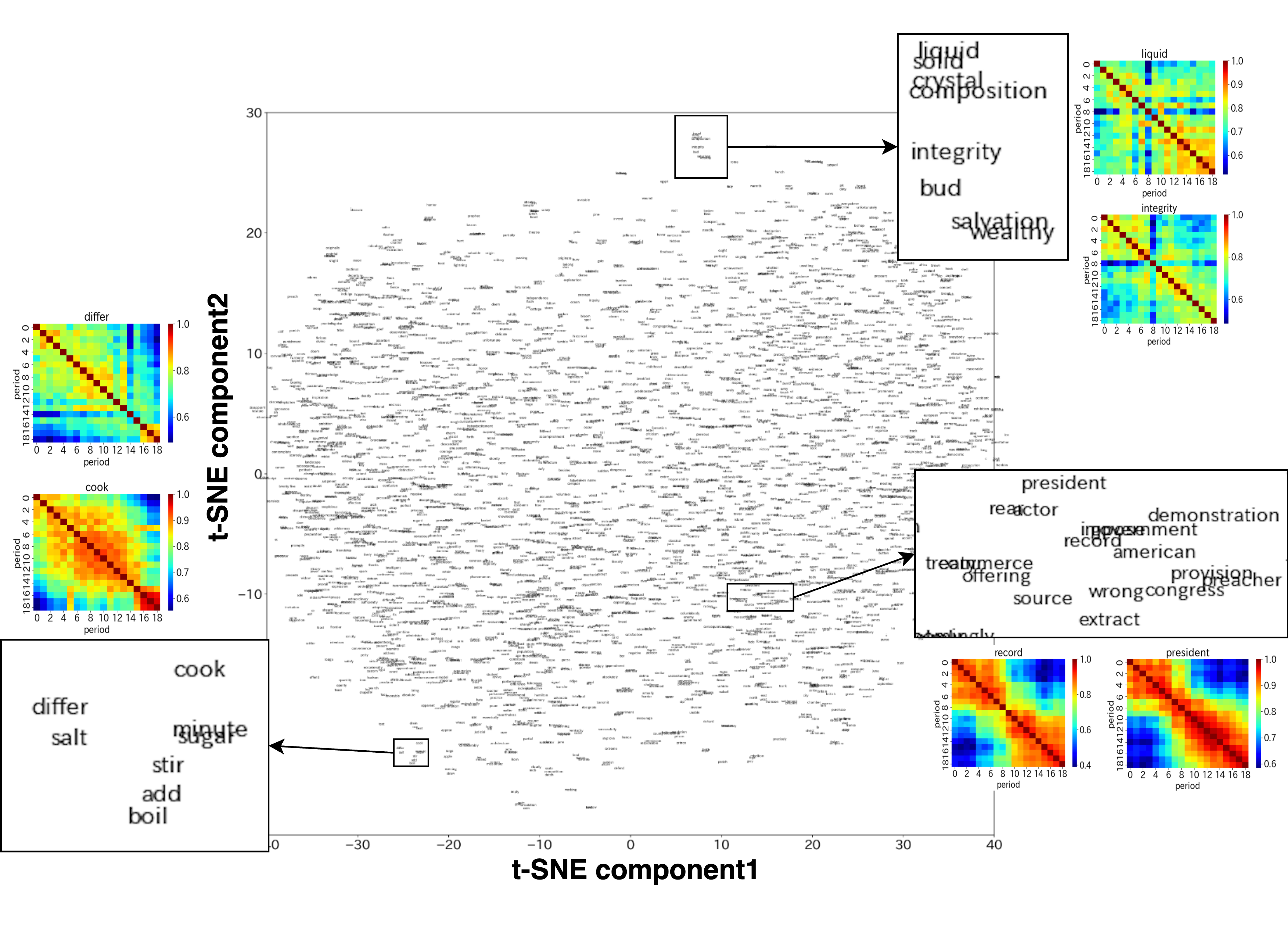}
    \caption{Visualizing the similarity matrix of all words in COHA using t-SNE in two dimensions shows that words close to each other in the compressed dimensions exhibit similar similarity patterns. }
    \label{fig:tsne-coha}
\end{figure*}

This analysis provides information that cannot be obtained by looking at only adjacent periods. 
When focusing solely on adjacent periods, we can observe a shift occurring at period 8 and another at period 27. 
However, in cases where the meaning reverts to its original state (as in the spike observed in period 8) or shifts to a completely different meaning (as seen in period 27), semantic shift analysis based only on adjacent periods can not distinguish these shifts. 
This ability to differentiate such cases is considered one of the advantages of analyzing semantic shifts across arbitrary time periods.

\subsubsection{Analysis of Diachronic Word Similarity Matrix by PPMI}
\label{subsec:simmat-ppmi}
Having identified the approximate change points for each word in \S \,\ref{subsec:vis-sim}, we conduct further investigation on driving factors of these shifts in each period.
From the diachronic word similarity matrix, we can manually interpret which periods exhibit high or low similarity by identifying regions of high similarity. 
By selecting appropriate periods from these regions and calculating the PPMI differences between them, we can conduct a deeper analysis of when semantic shifts occurred and how the co-occurring words changed over time.

To investigate whether the obtained similarity matrix reflects actual semantic shifts, we analyze it using the difference in PPMI between time periods. 
For the given target word $w$, we calculate the magnitude of the difference between the PPMI values between $t_1$ and $t_2$.
This difference $\Delta M^{t_2 \rightarrow t_1}$ measures the degree to which a context word $c$ co-occurs with $w$ in period $t_1$ but not in period $t_2$, and is used to extract the top-$k$ context words with the largest positive changes, defined as:
\begin{align*}
\Delta M^{(t_2 \rightarrow t_1)} &= M^{(t_1)}_{w*} - M^{(t_2)}_{w*}, \\
\mathcal{N}_k^{(t_2 \rightarrow t_1)} &= \{c \mid c \in \mathrm{argsort}(\Delta M^{(t_2 \rightarrow t_1)})_{[:k]}\}.
\end{align*}
Therefore, by calculating the difference and examining the top-ranked words $\mathcal{N}_k^{(t_2 \rightarrow t_1)}$ for the target word, it is possible to identify words that co-occur specifically in period $t_1$ and characterize the semantic shift intuitively.\footnote{Note that negative values in the PPMI differences between periods are not meaningful because of definition of PPMI.}

\begin{figure*}[t]
    \begin{tabular}{c}
    \centering
    \begin{minipage}{\columnwidth}
        \centering
        \includegraphics[scale=0.4]{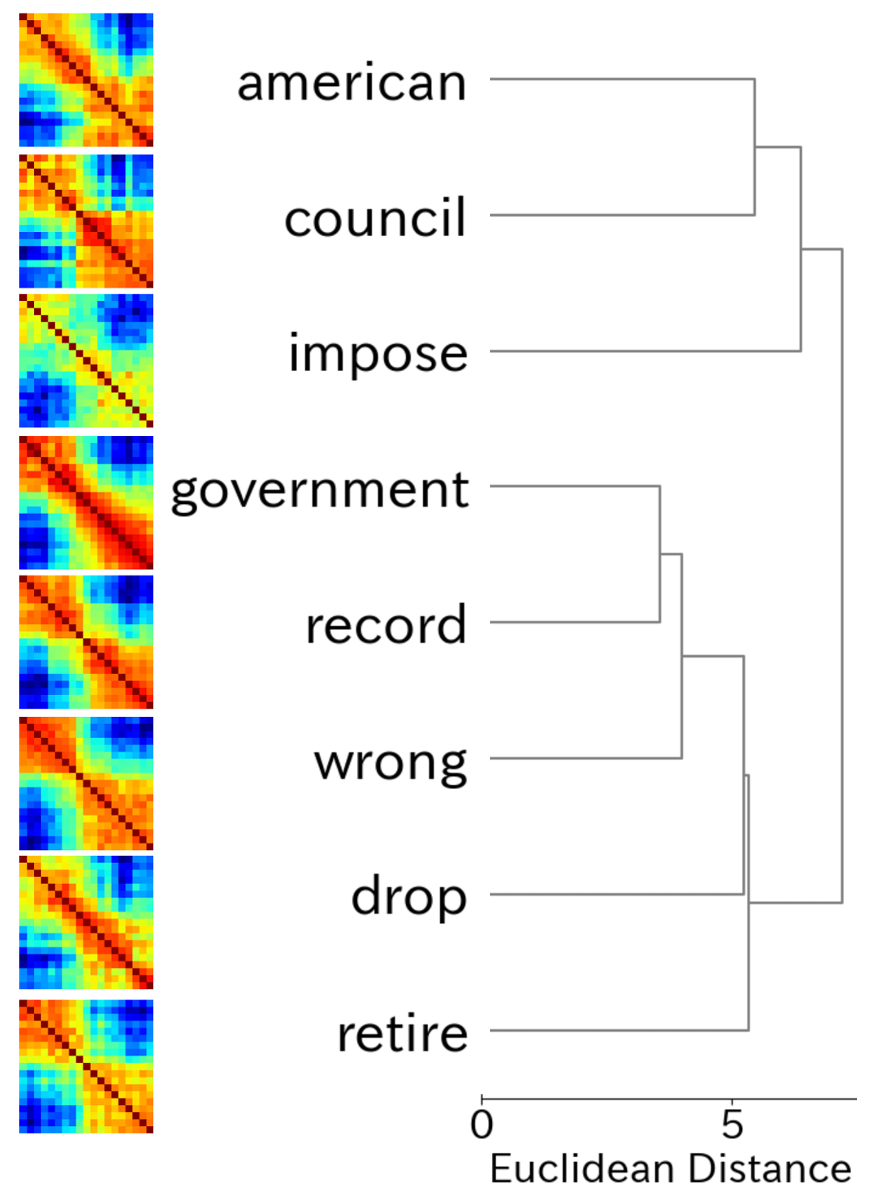}
        \subcaption{The results of the hierarchical clustering for words that are included in the same cluster as the word ``record'' in COHA.}
        \label{fig:coha-cluster}
    \end{minipage}
    \begin{minipage}{\columnwidth}
        \centering
        \includegraphics[scale=0.4]{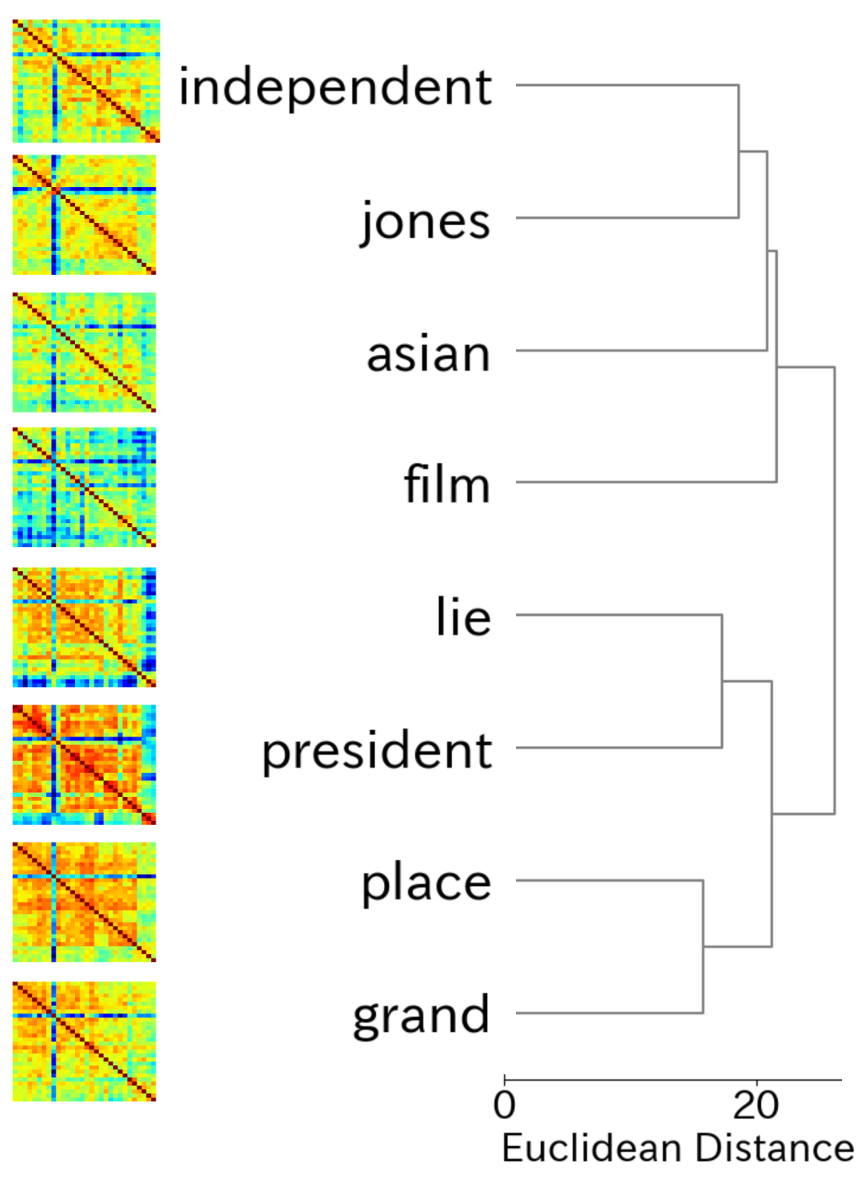}
        \subcaption{The results of the hierarchical clustering for words that are included in the same cluster as the word ``president'' in COCA.}
        \label{fig:coca-cluster}
    \end{minipage}
    \end{tabular}
    \caption{Visualization of clusters containing target words for each dataset was performed using hierarchical clustering for all words. This method allows us to observe how clusters that share a similar time-series pattern merge, providing insights into the clustering process and the relationships between words within the dataset.}
    \label{fig:cluster}
\end{figure*}

We check whether differences in similarity actually explain semantic shifts by examining the differences in PPMI ($\mathcal{N}_k^{(t_2 \rightarrow t_1)}$). 
\autoref{tab:record} shows the top 10 words with the greatest differences in PPMI for the word ``record'' between 1840 and 1940 in COHA.
The PPMI difference between 1840 and 1940 reveals co-occurring words specific to 1840, while subtracting the PPMI from 1940 from that of 1840 shows co-occurring words specific to 1940.
In 1840, the word ``record'' pertains to preserving and documenting events, with co-occurring words like ``chapter'' and ``journal'' reflecting this usage.
In 1940, the word ``record'' is used with meanings related to media for capturing sound, as evidenced by co-occurring words such as ``concert'' and ``sale''.

\autoref{tab:president} shows the top 10 words with the greatest differences in PPMI for the word ``president'' between 1991, 1998, 2012 and 2019 in COCA.
In 1991, words such as ``republic'', ``marketing'', and ``peace'' were commonly associated with the term ``president'', reflecting its use in relation to general policies and actions typically undertaken by a president.
In 1998, words like  ``video'',  ``mexican'', and  ``andrew'' appear. These terms suggest a connection to the documentary Presidential Train, which may be related to former U.S. President Andrew Jackson. 
In 2012, terms like ``ohio'', ``tax'', and ``immigration'' were identified, which can be seen as reflecting key issues in the 2012 presidential election.
In 2017, words such as ``border'' and ``moon'' emerged, likely linked to the priorities of the Trump administration, which began that year.
These observations suggest that the term ``president'' has shifted in meaning in response to the political and social context, particularly during presidential elections and significant political events.

\subsection{Clustering of Diachronic Word Similarity Matrices}
\label{subsec:real-class}
We attempt to further analyze the detail of semantic shifts.
By visualizing the similarity matrices of all words using t-SNE, we confirmed that locally similar patterns can be identified.
By analyzing the similarity matrices through clustering, it is possible to identify words that exhibit similar behavior such as the periods in which similarity shifts, the magnitude of the shift, and the periods where spikes occur. 
Our method allows for semantic shift analysis regardless of the period being studied. 
Among clusters, there are some that have shifted due to sociological factors, making it a technique that can be applied to trend analysis.

\begin{figure*}[t]
    \centering
    \includegraphics[scale=0.33]{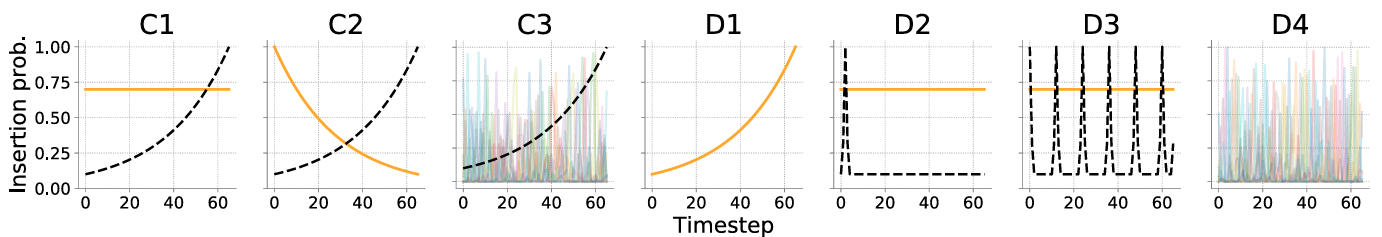}
    \caption{Illustration of seven schemas for inserting pseudowords into the synthetic dataset~\cite{shoemark-etal-2019-room}. The orange line represents $sense_1$, the black dotted line represents $sense_2$, and the other lines correspond to the remaining $senses$.}
    \label{fig:schema}
\end{figure*}

We present the results of visualizing the similarity matrix of all words using t-SNE, showing the obtained patterns and their distribution.
In the visualization of the overall similarity matrix, t-SNE\footnote{We used TSNE from scikit-learn.} was employed to reduce the dimensionality to two dimensions.
\autoref{fig:tsne-coha} displays the two-dimensional visualization of the similarity matrix for all words in COHA using t-SNE.\footnote{The result of COCA is provided in \autoref{sec:appendix:coca}}
In the similarity matrix compressed by t-SNE, words located at nearby coordinates exhibit similar patterns in their similarity matrices.
In \autoref{fig:tsne-coha}, we observed patterns such as the word ``record'', which shows separate regions of high similarity during intermediate periods, and the word ``liquid'', which exhibits a spike in similarity during specific periods, forming localized clusters.
Our study primarily aims to identify and group locally similar patterns of semantic shifts in an unsupervised manner. 
However, the t-SNE visualization reveals the potential to explore global patterns as well, which could provide valuable insights. This intriguing direction will be explored further as part of future work.

We present the results of clustering all the word similarity matrices using COHA and COCA.
In \autoref{fig:coha-cluster}, the results of the hierarchical clustering for words that are included in the same cluster as the word ``record'' in COHA are presented.
It is observed that ``record'' is clustered with words like ``government'' and ``wrong''. 
The words included in this cluster exhibit similarity matrices similar to that of ``record'', indicating the potential occurrence of a semantic shift. 
While it is necessary to analyze the PPMI differences to confirm whether a semantic shift has indeed occurred, \autoref{fig:coha-cluster} shows that this type of analysis is valuable for identifying candidate words that may have undergone semantic shifts.

In \autoref{fig:coca-cluster}, the results of the hierarchical clustering for words included in the same cluster as the word ``president'' in COCA are shown.
It is found that ``president'' is clustered with words like ``lie'' and ``independent''.
The words in this cluster share a similar similarity matrix with ``president'', suggesting that they may have experienced a semantic shift due to the same factors. A notable commonality among these clusters is the spike observed in period 8 (1998), which may be attributed to the release of a documentary related to former U.S. President Andrew Jackson. \autoref{fig:coca-cluster} shows that this highlights the ability of the analysis to detect clusters potentially influenced by social factors, enabling further exploration of semantic shifts driven by external events.

\section{Experiment: Pseudo Data}
\label{chap:pseudo}
To quantitatively demonstrate the validity of the proposed framework, we evaluate the classification performance of the seven pseudo-shift schemas proposed by \citet{shoemark-etal-2019-room}.\footnote{The experiments with pseudo data are conducted to find the optimal experimental settings for clustering in real data. 
We chose the classification task with seven shift schemas because it defines types of shifts in semantic shift research.}
First, we describe the experimental setup and the seven pseudo-shift schemas in \autoref{subsec:pseuod-set}.\footnote{Following by an analysis of the similarity matrices for each of the seven schemas in \autoref{subsec:exp-pseudo_simmat}. } 
Next, we present the results of the classification task for the seven schemas, demonstrating that using the similarity data across all time periods leads to better classification performance in \autoref{subsec:pseudo-class}.

\subsection{Experimental Setup}
\label{subsec:pseuod-set}

We classify the seven pseudo-shift schemas proposed by  \cite{shoemark-etal-2019-room} in \autoref{fig:schema}. 
They define three semantic shift schemas (C1-C3) and four non-semantic shift schemas (D1-D4). 
For each schema's pseudoword, two words extracted from the corpus, $word_1$ and $word_2$, are considered the \textbf{primary senses} of the pseudoword (each representing $sense_1$ and $sense_2$), while seven randomly extracted words $words \in \{word_3, ..., word_9\}$ are considered \textbf{miscellaneous senses} ($senses$). 
These words are then replaced with the pseudoword to reproduce the seven schemas.\footnote{Detailed experimental settings are provided in ~\autoref{sec:appendix:pseudo-setting}.}

\begin{description}[noitemsep]
\item \textbf{C1}: $sense_1$ remains constant while $sense_2$ increases over time (acquisition of a new sense)
\item \textbf{C2}: $sense_1$ decreases over time while $sense_2$ increases (sense transition)
\item \textbf{C3}: $sense_1$ increases over time, with seven $senses$ randomly selected each period (acquisition of noisy senses)
\item \textbf{D1}: $sense_1$ increases over time (increase of a sense)
\item \textbf{D2}: $sense_1$ remains constant while $sense_2$ spikes once (sensitive to a specific period)
\item \textbf{D3}: $sense_1$ remains constant while $sense_2$ spikes periodically (periodically sensitive shifts)
\item \textbf{D4}: Seven $senses$ are randomly selected each period (pure noise)
\end{description}

We create pseudo-data using the year of 2010 in  Mainichi Shimbun dataset.\footnote{\url{https://mainichi.jp/contents/edu/03.html}}
The dataset is sampled at 70\%, and 20 periods are created. 
Japanese texts are tokenized by  MeCab\footnote{\url{https://taku910.github.io/mecab/}} with UniDic\footnote{\url{https://clrd.ninjal.ac.jp/unidic/}}.
Words that appear 50 times or more are used as target words. 
The frequency set is divided into four quantiles, with five pseudowords prepared for each frequency set. 
For each schema, 20 pseudowords are prepared, resulting in a total of 140 pseudowords. 
We used a 100-dimensional PPMI-SVD joint~\cite{aida-etal-2021-comprehensive} for word embeddings.

The classification performance is evaluated by varying the similarity methods, feature extraction methods, clustering techniques and whether to apply normalization.
The input for clustering is the similarity matrices of the 140 pseudowords. 
The classification is performed with seven clusters, and the optimal label-cluster correspondence is obtained using the linear sum assignment algorithm\footnote{The calculation was performed using the linear\_sum\_assignment function from SciPy.} to calculate accuracy. 

We investigated the method for calculating similarity, feature extraction methods, clustering methods, and whether to apply normalization.
Cosine similarity and Euclidean distance are used as similarity measures. 
Three feature extraction methods of the similarity matrix are examined:
\textbf{Adjacent periods} using $S_{i(i+1)}(w)$ elements of the similarity matrix $S(w)$ of the similarity matrix, 
\textbf{Similarity with period 0} using $S_{0*}(w)$ elements of the similarity matrix $S(w)$,
\textbf{Upper triangular} components using $S_{ij}(w)$ elements of the similarity matrix $S(w)$, where $i < j$.
We investigate hierarchical clustering and K-means++.\footnote{We used AgglomerativeClustering and kmeans from scikit-learn. DBSCAN was not used because it cannot specify the number of clusters for classifying into seven clusters.}
The impact of normalization is also investigated.

\begin{table}[t]
\small
\centering
\begin{tabular}{lllll} \hline
\multirow{2}{*}{similarity} & \multirow{2}{*}{feature} & \multirow{2}{*}{cluster} & \multicolumn{2}{c}{accuracy} \\
&                &    & raw          & + stand        
\\ \hline 
\multirow{6}{*}{\begin{tabular}[c]{@{}l@{}}Cosine \\ similarity\end{tabular}}  & \multirow{2}{*}{Adjacent}& Agglo     & 53.6         & 70.7          \\
&  & K-means& 54.3 & 68.6 \\
 & \multirow{2}{*}{Period 0}& Agglo & 50.0   & 63.6  \\
  &      & K-means& 49.3         & 62.9          \\
  & \multirow{2}{*}{Upper Tri}& Agglo     & 55.7         & \textbf{72.1}          \\
&    & K-means& \textbf{57.9} & 71.4          \\ \hline
\multirow{6}{*}{\begin{tabular}[c]{@{}l@{}}Euclidean \\ distance\end{tabular}} & \multirow{2}{*}{Adjacent}& Agglo     & 32.8         & 56.4          \\
&                                  & K-means& 34.2         & 55.7          \\
    & \multirow{2}{*}{Period 0}          & Agglo     & 48.5         & 64.9          \\
&                                  & K-means& 47.8         & 66.4          \\
   & \multirow{2}{*}{Upper Tri}& Agglo     & 47.8         & 71.4          \\
&                                  & K-means& 45.7         & 68.5         \\ \hline
\end{tabular}
\caption{Classification performance for shift schemas in pseudo-data. 
The "Agglo" means hierarchical clustering. 
The "+ stand" indicates the accuracy achieved when the features were standardized.
}
\label{tab:pseudo}
\end{table}

\subsection{Classification of Shift Schemas}
\label{subsec:pseudo-class}
We discuss the results of classifying shift schemas under various experimental settings.
\autoref{tab:pseudo} presents the classification performance results for each method. 
\paragraph{Similarity.}
Regarding the similarity calculation methods, cosine similarity generally performs better. 
It is evident that normalization of norms is important for classifying shifts.
\paragraph{Features.}
The number of features is highest for the upper triangular elements, indicating that similarity between distant time periods is useful for classifying shift schemas.
\paragraph{Clustering methods.}
Concerning clustering methods, before standardization, K-means++ performs slightly better, but after standardization, hierarchical clustering shows a slight improvement in performance. 
When using cosine similarity, there is no significant difference in performance between K-means++ and hierarchical clustering.
\paragraph{Standardization.}
Additionally, comparing cases with and without standardization, it was found that performance improves with standardization. Since the task involves classifying shift schemas, capturing the movement of similarity rather than the similarity itself is beneficial.

\vskip \baselineskip


We conducted experiments on real data in \autoref{chap:simmat} using the combination (cosine similarity + upper triangular matrix  + hierarchical clustering + standardization) that achieved the highest classification  performance in the above experiments.\footnote{An error analysis is conducted on the best-performing results and confusion matrices of various cases in \autoref{sec:appendix:confusion}. The results of the t-SNE visualization for the pseudo-data are presented in \autoref{sec:appendix:vis-pseudo}}

\section{Conclusion}
We proposed a framework that enables analyzing and clustering semantic shifts across arbitrary periods shifts using diachronic word similarity matrices. 
The experiments with real data showed that the similarity matrices enables semantic shifts analysis across arbitrary time periods, and their clustering allows for unsupervised grouping of words with similar behaviors.
Additionally, the experiments with pseudo-data demonstrated that the proposed framework is well-suited for classifying shift schemas. 

We hope that this study will advance research on semantic shift across multiple time periods. By tracking how the meanings of words shift, it will become possible to conduct more detailed analyses of semantic shift phenomena, such as classifying patterns of semantic shifts using embeddings.

\section{Limitations}
\paragraph{Dataset Limitation}
We analyzed corpora that contain data from multiple time periods within the same domain.  
However, when dealing with multiple time periods but across different domains, the domain differences may be reflected in the similarity matrix, making it difficult to analyze semantic shifts. Therefore, developing robust methods that can be applied even in cases where the domains differ will be essential for future research on semantic shifts.
Additionally, for time period segmentation, we used a 10-year interval for COHA and a 1-year interval for COCA. 
Analyzing data in smaller time units, such as 1-month interval analysis with the NOW corpus\footnote{\url{https://www.english-corpora.org/now/}}, is a potential future direction. As the time units become smaller, the focus shifts from large-scale shifts to more subtle ones. 
Therefore, it will be important to investigate whether the analysis remains effective when adjusting the time unit.
However, it is important to note that the optimal time slices may vary depending on the analysis target, making evaluation challenging.

\paragraph{Embedding methods Limitation}
Only the PPMI-SVD joint embedding method was tested, which may cause a limitation of the embedding method itself.
If BERT-based embeddings~\cite{cassotti-etal-2023-xl,aida2024semanticdistancemetriclearning,periti2024analyzingsemanticchangelexical} were used for analysis, it would be possible to obtain word embeddings regardless of the dataset size.
In this study, we interpret the similarity matrices using differences in PPMI. However, in the case of dynamic embeddings, where word embeddings for each period are available, it is believed that interpretations can also be drawn from clustering results across different periods. 
However, BERT-based approach would require narrowing down the target words due to the computational demands of generating word embeddings.
In this study, thanks to computationally efficient PPMI-SVD, we were able to analyze a large set of target words.
When it comes to classifying similarity matrices, it is challenging to draw a clear line between which words to include and which to exclude.

\paragraph{Pseudo schemas Limitation}
We evaluated a classification task using pseudo data based on the seven shift schemas proposed by \citet{shoemark-etal-2019-room}. 
However, these seven schemas do not necessarily cover all types of semantic shifts. 
In our analysis of similarity matrices for real data, such as for the words ``record'' and ``president'', we observe shifts that do not fit into any of the defined schemas. 
Additionally, none of the schemas account for information across arbitrary time periods, which may lead to an underestimation of the proposed method's potential. 
As a future direction, it is worth expanding the definition of shift schemas that more closely reflect real data, specifically by formalizing the task of semantic shift across multiple periods.

\paragraph{Application Limitation}
An application in linguistics is the automatic identification of words that have undergone semantic shifts~\cite{cook-stevenson-2010-automatically}.
By leveraging computational methods to identify such words, linguists can prioritize them for analysis, enabling efficient exploration of newly shifted meanings.
Using this framework not only makes it possible to detect semantic shifts but also allows for multi-period analysis of how the shifts occurred. 
However, two issues warrant discussion: (i) how to select the target words for analysis and (ii) the fact that changes in similarity do not always correspond to semantic shifts. 
Some words exhibit behaviors that do not align with traditionally shifted words, making it necessary—but challenging—to define the degree of semantic shifts over multiple periods. 
Additionally, even when similarity changes, as shown in \autoref{fig:cos_all}(b), there are cases where the word's meaning has not shifted.
Addressing these challenges will not only improve the precision of automatic methods but also enhance their applicability to broader linguistic studies, paving the way for deeper insights into semantic shifts.

\section*{Ethical Consideration}
While this study does not involve creating or publishing new data or models, and thus no direct ethical concerns are anticipated, it is important to acknowledge that the publicly available corpora used for training word vectors may contain inherent biases. 
Additionally, the proposed method does not specify particular word vectors for constructing the input similarity matrices. However, when using pre-trained word vectors or masked language models like BERT, it is crucial to be aware of the potential biases these models might contain, which could influence the results~\cite{anantaprayoon-etal-2024-evaluating}. 
Addressing these biases is necessary to maintain the integrity and fairness of the research outcomes.

\section*{Acknowledgements}
This research was supported by the NINJAL collaborative research project and NINJAL Diachronic Corpus project at the National Institute for Japanese Language and Linguistics, Japan.
Also, this work was partly supported by JST, PRESTO Grant Number JPMJPR2366, Japan.

\bibliography{custom}

\begin{thebibliography}{41}
\providecommand{\natexlab}[1]{#1}

\bibitem[{Aida and Bollegala(2024)}]{aida2024semanticdistancemetriclearning}
Taichi Aida and Danushka Bollegala. 2024.
\newblock \href {https://aclanthology.org/2024.findings-acl.451} {A semantic distance metric learning approach for lexical semantic change detection}.
\newblock In \emph{ACL findings 2024}, pages 7570--7584, Bangkok, Thailand and virtual meeting. Association for Computational Linguistics.

\bibitem[{Aida et~al.(2021)Aida, Komachi, Ogiso, Takamura, and Mochihashi}]{aida-etal-2021-comprehensive}
Taichi Aida, Mamoru Komachi, Toshinobu Ogiso, Hiroya Takamura, and Daichi Mochihashi. 2021.
\newblock \href {https://aclanthology.org/2021.paclic-1.3} {A comprehensive analysis of {PMI}-based models for measuring semantic differences}.
\newblock In \emph{PACLIC 2021}, pages 21--31, Shanghai, China. Association for Computational Lingustics.

\bibitem[{Alatrash et~al.(2020)Alatrash, Schlechtweg, Kuhn, and Schulte~im Walde}]{alatrash-etal-2020-ccoha}
Reem Alatrash, Dominik Schlechtweg, Jonas Kuhn, and Sabine Schulte~im Walde. 2020.
\newblock \href {https://aclanthology.org/2020.lrec-1.859} {{CCOHA}: Clean corpus of historical {A}merican {E}nglish}.
\newblock In \emph{LREC 2020}, pages 6958--6966, Marseille, France. European Language Resources Association.

\bibitem[{Anantaprayoon et~al.(2024)Anantaprayoon, Kaneko, and Okazaki}]{anantaprayoon-etal-2024-evaluating}
Panatchakorn Anantaprayoon, Masahiro Kaneko, and Naoaki Okazaki. 2024.
\newblock \href {https://aclanthology.org/2024.lrec-main.566} {Evaluating gender bias of pre-trained language models in natural language inference by considering all labels}.
\newblock In \emph{LREC-COLING 2024}, pages 6395--6408, Torino, Italia. ELRA and ICCL.

\bibitem[{Baes et~al.(2024)Baes, Haslam, and Vylomova}]{baes-etal-2024-multidimensional}
Naomi Baes, Nick Haslam, and Ekaterina Vylomova. 2024.
\newblock \href {https://aclanthology.org/2024.acl-long.76} {A multidimensional framework for evaluating lexical semantic change with social science applications}.
\newblock In \emph{ACL 2024}, pages 1390--1415, Bangkok, Thailand. Association for Computational Linguistics.

\bibitem[{Cassotti et~al.(2024)Cassotti, De~Pascale, and Tahmasebi}]{cassotti2024usingsynchronicdefinitionssemantic}
Pierluigi Cassotti, Stefano De~Pascale, and Nina Tahmasebi. 2024.
\newblock \href {https://aclanthology.org/2024.acl-long.249} {Using synchronic definitions and semantic relations to classify semantic change types}.
\newblock In \emph{ACL 2024}, pages 4539--4553, Bangkok, Thailand. Association for Computational Linguistics.

\bibitem[{Cassotti et~al.(2023)Cassotti, Siciliani, DeGemmis, Semeraro, and Basile}]{cassotti-etal-2023-xl}
Pierluigi Cassotti, Lucia Siciliani, Marco DeGemmis, Giovanni Semeraro, and Pierpaolo Basile. 2023.
\newblock \href {https://doi.org/10.18653/v1/2023.acl-short.135} {{XL}-{LEXEME}: {W}i{C} pretrained model for cross-lingual {LEX}ical s{EM}antic chang{E}}.
\newblock In \emph{ACL 2023}, pages 1577--1585, Toronto, Canada.

\bibitem[{Cook and Stevenson(2010)}]{cook-stevenson-2010-automatically}
Paul Cook and Suzanne Stevenson. 2010.
\newblock \href {http://www.lrec-conf.org/proceedings/lrec2010/pdf/657_Paper.pdf} {Automatically identifying changes in the semantic orientation of words}.
\newblock In \emph{LREC 2010}, Valletta, Malta.

\bibitem[{Davies(2009)}]{coca}
Mark Davies. 2009.
\newblock The 385+ million word corpus of contemporary american english (1990--2008+): Design, architecture, and linguistic insights.
\newblock \emph{International journal of corpus linguistics}, 14(2):159--190.

\bibitem[{Davies(2012)}]{coha}
Mark Davies. 2012.
\newblock Expanding horizons in historical linguistics with the 400-million word corpus of historical american english.
\newblock \emph{Corpora}, 7(2):121--157.

\bibitem[{de~Sá et~al.(2024)de~Sá, Silveira, and Pruski}]{desa-2024-surveycharacterizationsemanticchange}
Jader Martins~Camboim de~Sá, Marcos~Da Silveira, and Cédric Pruski. 2024.
\newblock \href {https://arxiv.org/abs/2402.19088} {Survey in characterization of semantic change}.
\newblock \emph{Preprint}, arXiv:2402.19088.

\bibitem[{Firth(1957)}]{distributional-firth}
John~Rupert Firth. 1957.
\newblock A synopsis of linguistic theory.
\newblock \emph{In Studies in Linguistic Analysis (pp. 1-31). Special Volume of the Philological Society.}

\bibitem[{Giulianelli et~al.(2020)Giulianelli, Del~Tredici, and Fern{\'a}ndez}]{giulianelli-etal-2020-analysing}
Mario Giulianelli, Marco Del~Tredici, and Raquel Fern{\'a}ndez. 2020.
\newblock \href {https://doi.org/10.18653/v1/2020.acl-main.365} {Analysing lexical semantic change with contextualised word representations}.
\newblock In \emph{ACL 2020}, pages 3960--3973, Online.

\bibitem[{Hamilton et~al.(2016)Hamilton, Leskovec, and Jurafsky}]{hamilton-etal-2016-diachronic}
William~L. Hamilton, Jure Leskovec, and Dan Jurafsky. 2016.
\newblock \href {https://doi.org/10.18653/v1/P16-1141} {Diachronic word embeddings reveal statistical laws of semantic change}.
\newblock In \emph{ACL 2016}, pages 1489--1501, Berlin, Germany.

\bibitem[{Harris(1954)}]{distributional-harris}
Zellig~S. Harris. 1954.
\newblock Distributional structure.
\newblock \emph{WORD 10 (2–3)}.

\bibitem[{Hu et~al.(2019)Hu, Li, and Liang}]{hu-etal-2019-diachronic}
Renfen Hu, Shen Li, and Shichen Liang. 2019.
\newblock \href {https://doi.org/10.18653/v1/P19-1379} {Diachronic sense modeling with deep contextualized word embeddings: An ecological view}.
\newblock In \emph{ACL 2019}, pages 3899--3908, Florence, Italy. Association for Computational Linguistics.

\bibitem[{Inoue et~al.(2022)Inoue, Komachi, Ogiso, Takamura, and Mochihashi}]{inoue-etal-2022-infinite}
Seiichi Inoue, Mamoru Komachi, Toshinobu Ogiso, Hiroya Takamura, and Daichi Mochihashi. 2022.
\newblock \href {https://doi.org/10.18653/v1/2022.emnlp-main.104} {Infinite {SCAN}: An infinite model of diachronic semantic change}.
\newblock In \emph{EMNLP 2022}, pages 1605--1616, Abu Dhabi, United Arab Emirates. Association for Computational Linguistics.

\bibitem[{Ishihara et~al.(2022)Ishihara, Takahashi, and Shirai}]{ishihara-etal-2022-semantic}
Shotaro Ishihara, Hiromu Takahashi, and Hono Shirai. 2022.
\newblock \href {https://doi.org/10.18653/v1/2022.aacl-main.17} {Semantic shift stability: Efficient way to detect performance degradation of word embeddings and pre-trained language models}.
\newblock In \emph{AACL 2022}, pages 205--216, Online only. Association for Computational Linguistics.

\bibitem[{Kacem et~al.(2020)Kacem, Daoudi, Amor, Berretti, and Alvarez-Paiva}]{kacem-gram-matrix-3d}
A.~Kacem, M.~Daoudi, B.~Amor, S.~Berretti, and J.~Alvarez-Paiva. 2020.
\newblock \href {https://doi.org/10.1109/TPAMI.2018.2872564} {A novel geometric framework on gram matrix trajectories for human behavior understanding}.
\newblock \emph{IEEE Transactions on Pattern Analysis \& Machine Intelligence}, 42(01):1--14.

\bibitem[{Kulkarni et~al.(2015)Kulkarni, Al-Rfou, Perozzi, and Skiena}]{change-detection-2014}
Vivek Kulkarni, Rami Al-Rfou, Bryan Perozzi, and Steven Skiena. 2015.
\newblock \href {https://doi.org/10.1145/2736277.2741627} {Statistically significant detection of linguistic change}.
\newblock In \emph{WWW 2015}, page 625–635, Republic and Canton of Geneva, CHE. International World Wide Web Conferences Steering Committee.

\bibitem[{Kutuzov et~al.(2018)Kutuzov, {\O}vrelid, Szymanski, and Velldal}]{kutuzov-survey-2018}
Andrey Kutuzov, Lilja {\O}vrelid, Terrence Szymanski, and Erik Velldal. 2018.
\newblock \href {https://aclanthology.org/C18-1117} {Diachronic word embeddings and semantic shifts: a survey}.
\newblock In \emph{COLING 2018}, pages 1384--1397, Santa Fe, New Mexico, USA. Association for Computational Linguistics.

\bibitem[{Laicher et~al.(2021)Laicher, Kurtyigit, Schlechtweg, Kuhn, and Schulte~im Walde}]{laicher-etal-2021-explaining}
Severin Laicher, Sinan Kurtyigit, Dominik Schlechtweg, Jonas Kuhn, and Sabine Schulte~im Walde. 2021.
\newblock \href {https://doi.org/10.18653/v1/2021.eacl-srw.25} {Explaining and improving {BERT} performance on lexical semantic change detection}.
\newblock In \emph{EACL SRW 2021}, pages 192--202, Online. Association for Computational Linguistics.

\bibitem[{Lazaridou et~al.(2021)Lazaridou, Kuncoro, Gribovskaya, Agrawal, Liska, Terzi, Gimenez, de~Masson~d'Autume, Ko{\v{c}}isk{\'y}, Ruder, Yogatama, Cao, Young, and Blunsom}]{lazaridou-2021-mind}
Angeliki Lazaridou, Adhiguna Kuncoro, Elena Gribovskaya, Devang Agrawal, Adam Liska, Tayfun Terzi, Mai Gimenez, Cyprien de~Masson~d'Autume, Tom{\'a}{\v{s}} Ko{\v{c}}isk{\'y}, Sebastian Ruder, Dani Yogatama, Kris Cao, Susannah Young, and Phil Blunsom. 2021.
\newblock \href {https://openreview.net/forum?id=73OmmrCfSyy} {Mind the gap: Assessing temporal generalization in neural language models}.
\newblock In \emph{NeurIPS 2021}.

\bibitem[{Levy and Goldberg(2014)}]{levy-2014-matrix}
Omer Levy and Yoav Goldberg. 2014.
\newblock \href {https://proceedings.neurips.cc/paper_files/paper/2014/file/feab05aa91085b7a8012516bc3533958-Paper.pdf} {Neural word embedding as implicit matrix factorization}.
\newblock In \emph{NeurIPS 2014}, volume~27.

\bibitem[{Levy et~al.(2015)Levy, Goldberg, and Dagan}]{levy-etal-2015-improving}
Omer Levy, Yoav Goldberg, and Ido Dagan. 2015.
\newblock \href {https://doi.org/10.1162/tacl_a_00134} {Improving distributional similarity with lessons learned from word embeddings}.
\newblock \emph{Transactions of the Association for Computational Linguistics}, 3:211--225.

\bibitem[{Neto et~al.(2021)Neto, Pacheco, and Luvizon}]{antonio-gram-sound-2021}
Antonio~Joia Neto, Andre G.~C. Pacheco, and Diogo~Carbonera Luvizon. 2021.
\newblock \href {https://doi.org/10.1109/ICASSP39728.2021.9414168} {Improving deep learning sound events classifiers using gram matrix feature-wise correlations}.
\newblock In \emph{ICASSP 2021}, pages 3780--3784.

\bibitem[{Nylund et~al.(2024)Nylund, Gururangan, and Smith}]{nylund-etal-2024-time}
Kai Nylund, Suchin Gururangan, and Noah Smith. 2024.
\newblock \href {https://aclanthology.org/2024.acl-long.141} {Time is encoded in the weights of finetuned language models}.
\newblock In \emph{ACL 2024}, pages 2571--2587, Bangkok, Thailand. Association for Computational Linguistics.

\bibitem[{Periti et~al.(2024)Periti, Cassotti, Dubossarsky, and Tahmasebi}]{periti2024analyzingsemanticchangelexical}
Francesco Periti, Pierluigi Cassotti, Haim Dubossarsky, and Nina Tahmasebi. 2024.
\newblock \href {https://aclanthology.org/2024.acl-long.246} {Analyzing semantic change through lexical replacements}.
\newblock In \emph{ACL 2024}, pages 4495--4510, Bangkok, Thailand. Association for Computational Linguistics.

\bibitem[{Periti and Montanelli(2024)}]{Periti-survey-2024}
Francesco Periti and Stefano Montanelli. 2024.
\newblock \href {https://doi.org/10.1145/3672393} {Lexical semantic change through large language models: a survey}.
\newblock \emph{ACM Computing Surveys}, 56(11):1–38.

\bibitem[{Periti and Tahmasebi(2024{\natexlab{a}})}]{periti-tahmasebi-2024-systematic}
Francesco Periti and Nina Tahmasebi. 2024{\natexlab{a}}.
\newblock \href {https://doi.org/10.18653/v1/2024.naacl-long.240} {A systematic comparison of contextualized word embeddings for lexical semantic change}.
\newblock In \emph{NAACL 2024}, pages 4262--4282, Mexico City, Mexico. Association for Computational Linguistics.

\bibitem[{Periti and Tahmasebi(2024{\natexlab{b}})}]{periti-tahmasebi-2024-towards}
Francesco Periti and Nina Tahmasebi. 2024{\natexlab{b}}.
\newblock \href {https://aclanthology.org/2024.lchange-1.10} {Towards a complete solution to lexical semantic change: an extension to multiple time periods and diachronic word sense induction}.
\newblock In \emph{LChange 2024}, pages 108--119, Bangkok, Thailand. Association for Computational Linguistics.

\bibitem[{Qiu et~al.(2023)Qiu, Shi, Tan, Qu, Fang, Wang, Gao, Wu, and Li}]{qiu-gram-nystagmography}
Xihe Qiu, Shaojie Shi, Xiaoyu Tan, Chao Qu, Zhijun Fang, Hailing Wang, Yongbin Gao, Peixia Wu, and Huawei Li. 2023.
\newblock \href {https://doi.org/10.1109/ICCV51070.2023.01951} {Gram-based attentive neural ordinary differential equations network for video nystagmography classification}.
\newblock In \emph{ICCV 2023}, pages 21282--21291.

\bibitem[{Quentin et~al.(2017)Quentin, Benjamin, and Jean-Pierre}]{feltgen-freq-pattern-2017}
Feltgen Quentin, Fagard Benjamin, and Nadal Jean-Pierre. 2017.
\newblock \href {https://royalsocietypublishing.org/doi/10.1098/rsos.170830} {Frequency patterns of semantic change: corpus-based evidence of a near-critical dynamics in language change}.
\newblock \emph{Royal Society Open Science}.

\bibitem[{Rosin et~al.(2022)Rosin, Guy, and Radinsky}]{10.1145/3488560.3498529}
Guy~D. Rosin, Ido Guy, and Kira Radinsky. 2022.
\newblock \href {https://doi.org/10.1145/3488560.3498529} {Time masking for temporal language models}.
\newblock In \emph{WSDM 2022}, page 833–841, New York, NY, USA. Association for Computing Machinery.

\bibitem[{Rosin and Radinsky(2022)}]{rosin-radinsky-2022-temporal}
Guy~D. Rosin and Kira Radinsky. 2022.
\newblock \href {https://doi.org/10.18653/v1/2022.findings-naacl.112} {Temporal attention for language models}.
\newblock In \emph{NAACL Findings 2022}, pages 1498--1508, Seattle, United States.

\bibitem[{Schlechtweg et~al.(2020)Schlechtweg, McGillivray, Hengchen, Dubossarsky, and Tahmasebi}]{schlechtweg-etal-2020-semeval}
Dominik Schlechtweg, Barbara McGillivray, Simon Hengchen, Haim Dubossarsky, and Nina Tahmasebi. 2020.
\newblock \href {https://doi.org/10.18653/v1/2020.semeval-1.1} {{S}em{E}val-2020 task 1: Unsupervised lexical semantic change detection}.
\newblock In \emph{SemEval 2020}, pages 1--23, Barcelona (online). International Committee for Computational Linguistics.

\bibitem[{Shoemark et~al.(2019)Shoemark, Liza, Nguyen, Hale, and McGillivray}]{shoemark-etal-2019-room}
Philippa Shoemark, Farhana~Ferdousi Liza, Dong Nguyen, Scott Hale, and Barbara McGillivray. 2019.
\newblock \href {https://doi.org/10.18653/v1/D19-1007} {Room to {G}lo: A systematic comparison of semantic change detection approaches with word embeddings}.
\newblock In \emph{EMNLP-IJCNLP 2019}, pages 66--76, Hong Kong, China. Association for Computational Linguistics.

\bibitem[{Stern(1931)}]{stern-semantic-change}
Gustaf Stern. 1931.
\newblock \emph{Meaning and change of meaning: with special reference to the English language}.
\newblock Bloomington: Indiana University Press.

\bibitem[{Su et~al.(2022)Su, Tang, Guan, Wu, Zhang, and Li}]{su-etal-2022-improving}
Zhaochen Su, Zecheng Tang, Xinyan Guan, Lijun Wu, Min Zhang, and Juntao Li. 2022.
\newblock \href {https://doi.org/10.18653/v1/2022.emnlp-main.428} {Improving temporal generalization of pre-trained language models with lexical semantic change}.
\newblock In \emph{EMNLP 2022}, pages 6380--6393, Abu Dhabi, United Arab Emirates.

\bibitem[{Yao et~al.(2018)Yao, Sun, Ding, Rao, and Xiong}]{yao-evolving-2018}
Zijun Yao, Yifan Sun, Weicong Ding, Nikhil Rao, and Hui Xiong. 2018.
\newblock \href {https://doi.org/10.1145/3159652.3159703} {Dynamic word embeddings for evolving semantic discovery}.
\newblock In \emph{WSDM 2018}, page 673–681, New York, NY, USA. Association for Computing Machinery.

\bibitem[{Zhang et~al.(2016)Zhang, Wang, Gou, Sznaier, and Camps}]{zhang-gram-3d}
X.~Zhang, Y.~Wang, M.~Gou, M.~Sznaier, and O.~Camps. 2016.
\newblock \href {https://doi.org/10.1109/CVPR.2016.487} {Efficient temporal sequence comparison and classification using gram matrix embeddings on a riemannian manifold}.
\newblock In \emph{CVPR 2016}, pages 4498--4507, Los Alamitos, CA, USA. IEEE Computer Society.

\end{thebibliography}

\appendix

\section{Real Data in English}
\subsection{Setup}
\label{sec:appendix:real-setting}
We applied the CCOHA cleaning method ~\cite{alatrash-etal-2020-ccoha} to both COHA and COCA. 
For COHA, due to the dataset size, we used data starting from 1830, excluding the data from 1820. 
For COCA, web and blog data are not used because they do not include time-specific information. 
Afterward, we cleaned both COHA and COCA by removing symbols and performing subsampling. 
We used a subsampling threshold of $1e^{-4}$ for COHA and $e1^{-5}$ for COCA. 
These settings were chosen to ensure that target words with a frequency of 100 or more occurrences were not removed.
The statistics for COHA and COCA are shown in \autoref{tab:app-coha}, and \autoref{tab:app-coca}.

\begin{table}[t]
\small
\centering
\begin{tabular}{cccc}
\hline
Time Period & \#token & Time Period & \#token \\ \hline
1830        & 2,269,396   &1930        & 3,977,506   \\
1840        & 2,636,944   &1940        & 3,892,645   \\
1850        & 2,743,776   &1950        & 4,059,038   \\
1860        & 2,776,254   &1960        & 3,805,188  \\
1870        & 3,021,121   &1970        & 3,758,512   \\
1880        & 3,227,468   &1980        & 3,959,571   \\
1890        & 3,322,748   &1990        & 4,627,047   \\
1900        & 3,377,908   &2000        & 5,067,374   \\
1910        & 3,490,015   &2010        & 5,386,789   \\
1920        & 4,059,039   &- & -\\ \hline
\end{tabular}
\caption{Statistics of COHA.}
\label{tab:app-coha}
\end{table}

\begin{table}[t]
\small
\centering
\begin{tabular}{cccc}
\hline
Time Period & \#token &Time Period & \#token \\ \hline
1990        & 1,912,238   &2005        & 1,991,792   \\
1991        & 1,934,292   &2006        & 1,994,816   \\
1992        & 1,895,978   &2007        & 1,942,283   \\
1993        & 1,902,687   &2008        & 1,922,213   \\
1994        & 1,947,562   &2009        & 1,876,034   \\
1995        & 1,949,698   &2010        & 1,798,676   \\
1996        & 1,917,884   &2011        & 2,008,111   \\
1997        & 1,942,688   &2012        & 2,042,931   \\
1998        & 1,951,724   &2013        & 1,928,037   \\
1999        & 1,941,432   &2014        & 1,918,909   \\
2000        & 1,993,549   &2015        & 1,942,229   \\
2001        & 1,916,218   &2016        & 1,927,982   \\
2002        & 1,974,183   &2017        & 1,998,353   \\
2003        & 1,995,176   &2018        & 2,019,840   \\
2004        & 1,993,568   &2019        & 2,030,788   \\ \hline
\end{tabular}
\caption{Statistics of COCA.}
\label{tab:app-coca}
\end{table}

\begin{table}[t]
\small
\centering
\begin{tabular}{cccc}
\hline
Time Period & \#token & Time Period & \#token \\ \hline
2003        & 5,590,218  &2012        & 5,300,054   \\
2004        & 5,159,294  &2013        & 5,055,583   \\
2005        & 4,953,488   &2014        & 5,400,230   \\
2006        & 4,659,049   &2015        & 5,302,944   \\
2007        & 4,538,968   &2016        & 5,202,454   \\
2008        & 4,314,940   &2017        & 5,161,849   \\
2009        & 4,294,827   &2018        & 4,827,454   \\
2010        & 4,288,985   &2019        & 4,302,899   \\
2011        & 4,553,186   &2020        & 3,786,987   \\ \hline
\end{tabular}
\caption{Statistics of Mainichi Shimbun dataset.}
\label{tab:app-mai}
\end{table}

\subsection{Silhouette Score}
\label{sec:appendix:class}
\begin{figure*}[t]
        \centering
    \begin{minipage}[b]{0.45\textwidth}
        \centering
        \includegraphics[scale=0.18]{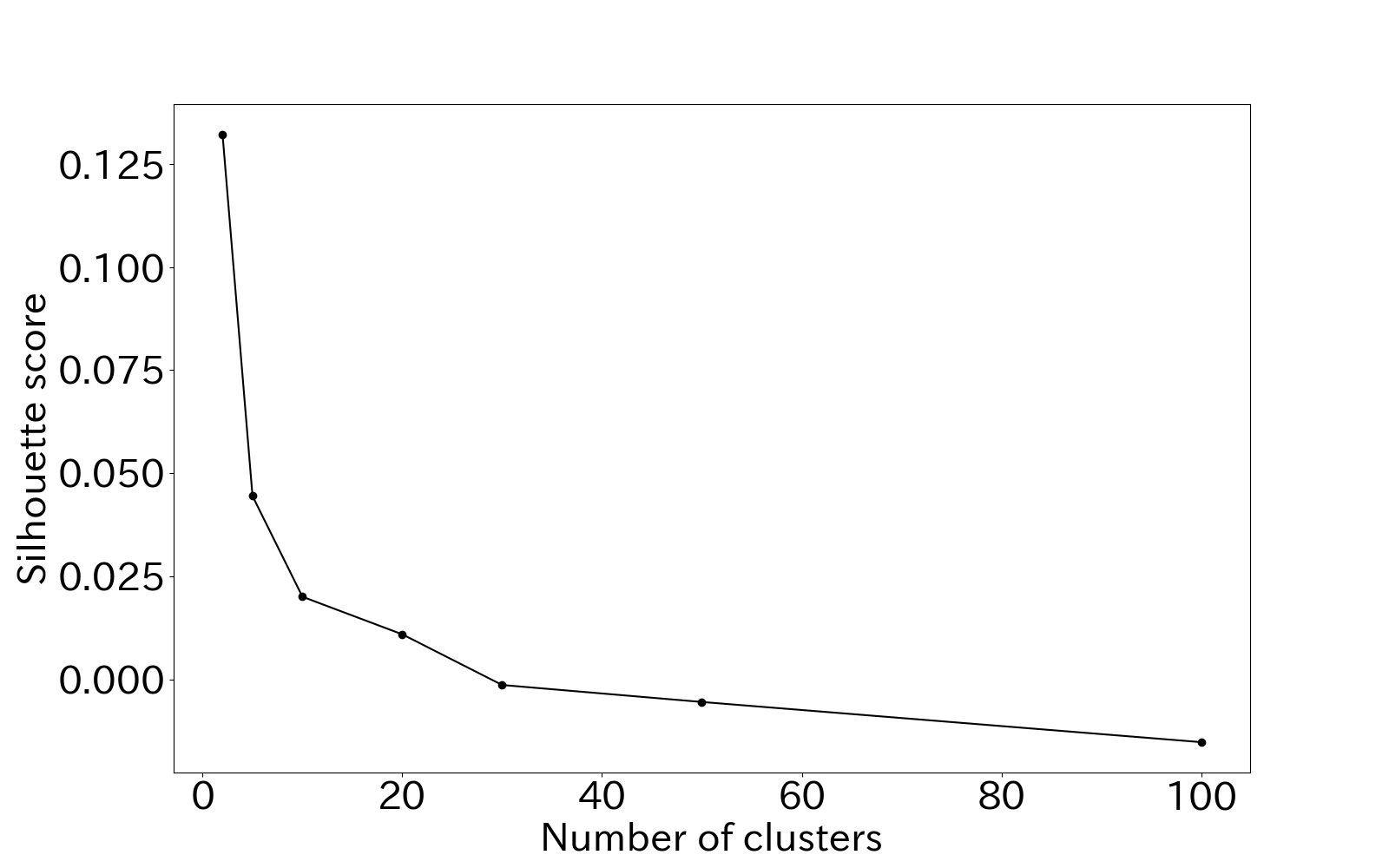}
        \subcaption{K-means+COHA}
        \label{fig:sil_coha_k}
    \end{minipage}
    \begin{minipage}[b]{0.45\textwidth}
        \centering
        \includegraphics[scale=0.18]{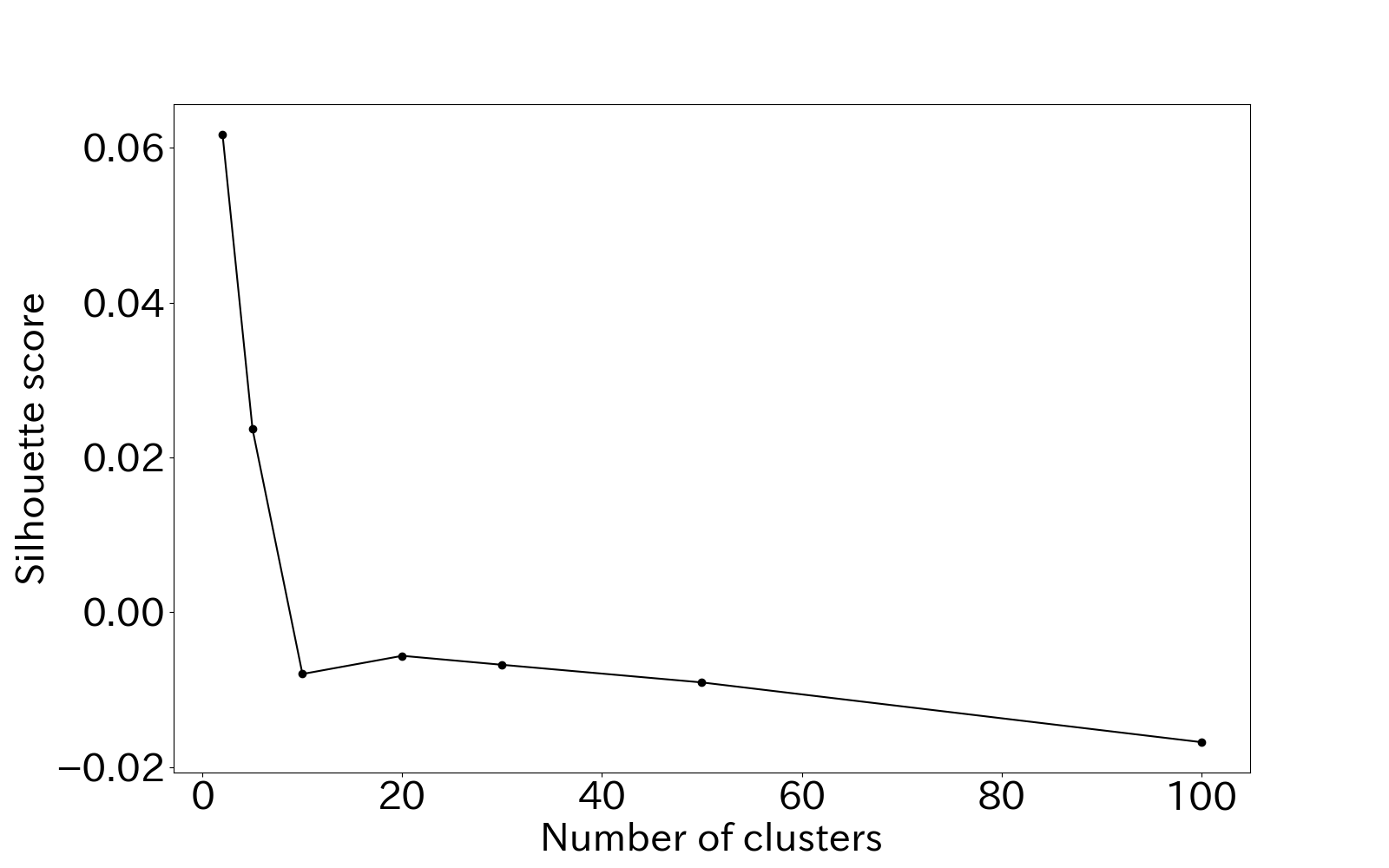}
        \subcaption{K-means+COCA}
        \label{fig:sil_coca_k}
    \end{minipage}
    \\
    \begin{minipage}[b]{0.45\textwidth}
        \centering
        \includegraphics[scale=0.18]{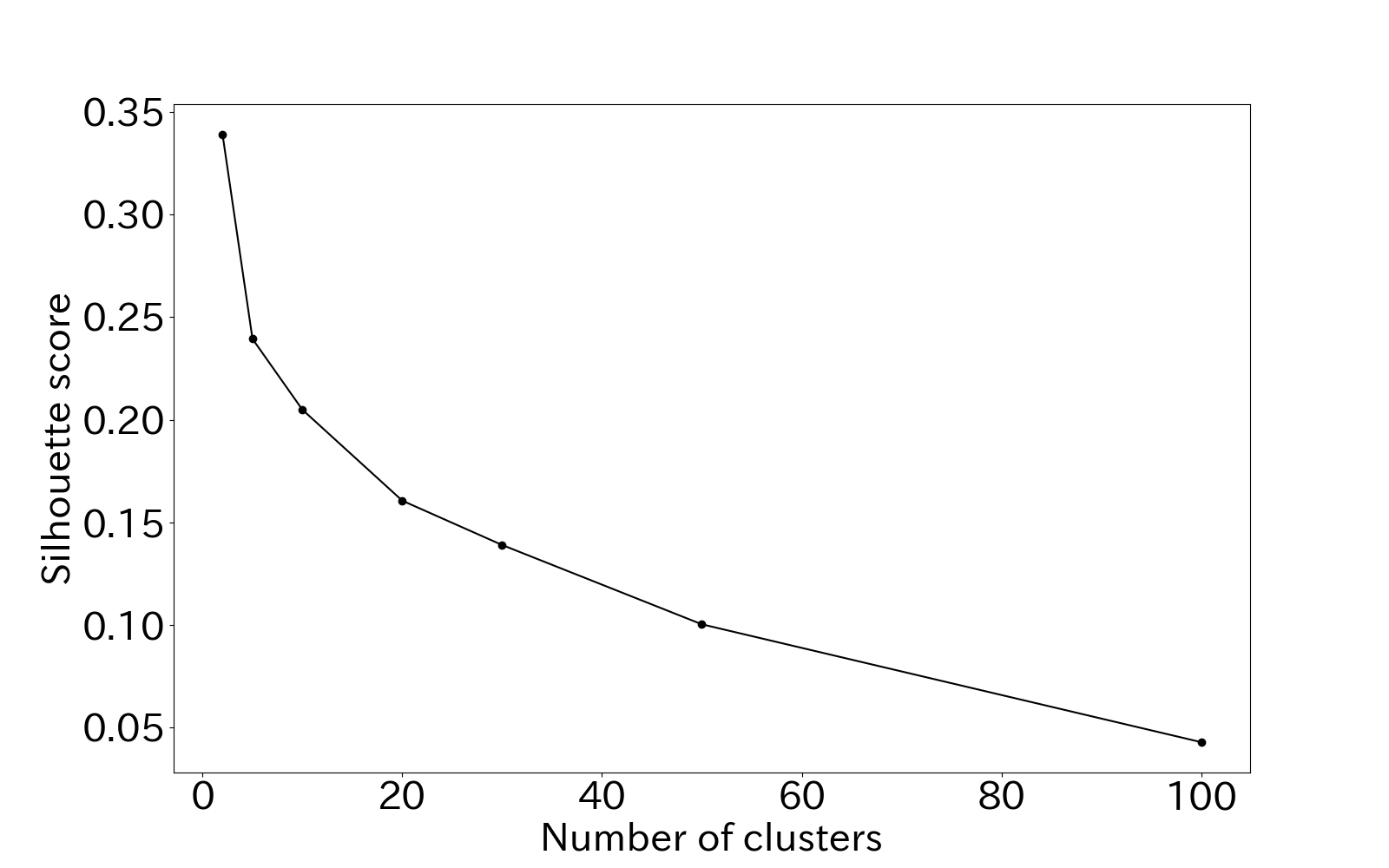}
        \subcaption{Agglo+COHA}
        \label{fig:sil_coha_a}
    \end{minipage}
    \begin{minipage}[b]{0.45\textwidth}
        \centering
        \includegraphics[scale=0.18]{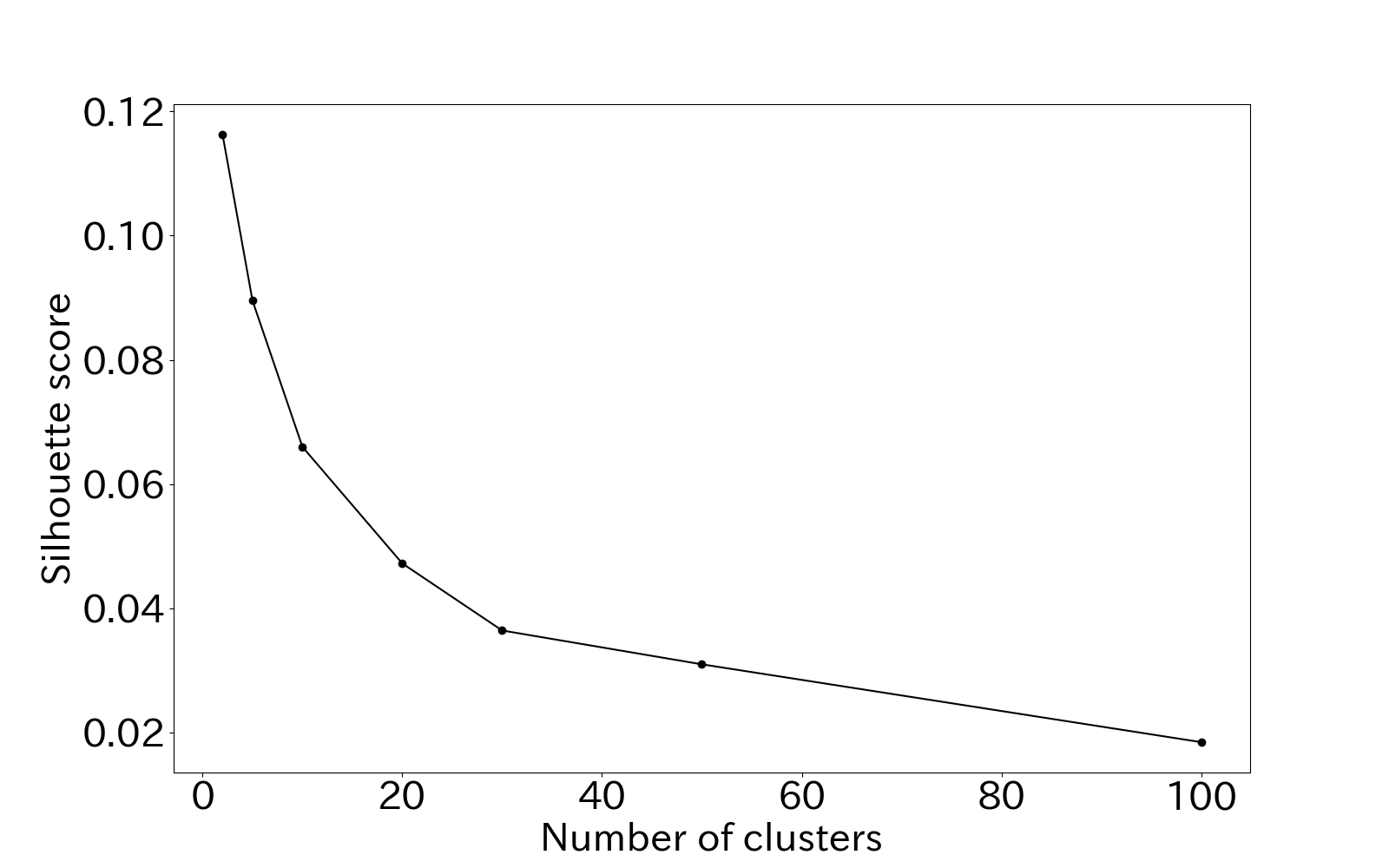}
        \subcaption{Agglo+COCA}
        \label{fig:sil_coca_a}
    \end{minipage}
    \caption{Silhouette score of each clustering method in each dataset. Agglo means hierarchical clustering.}
    \label{fig:sil}
\end{figure*}

In the clustering process, the optimal number of clusters was determined using the silhouette score. The progression of silhouette scores for both COHA and COCA datasets, using K-means and hierarchical clustering, is shown in \autoref{fig:sil}. The results consistently indicate that the optimal number of clusters is 2 across all cases.

\subsection{t-SNE Visualization in COCA}
\label{sec:appendix:coca}
\begin{figure*}[t]
    \centering
    \includegraphics[keepaspectratio, scale=0.095]{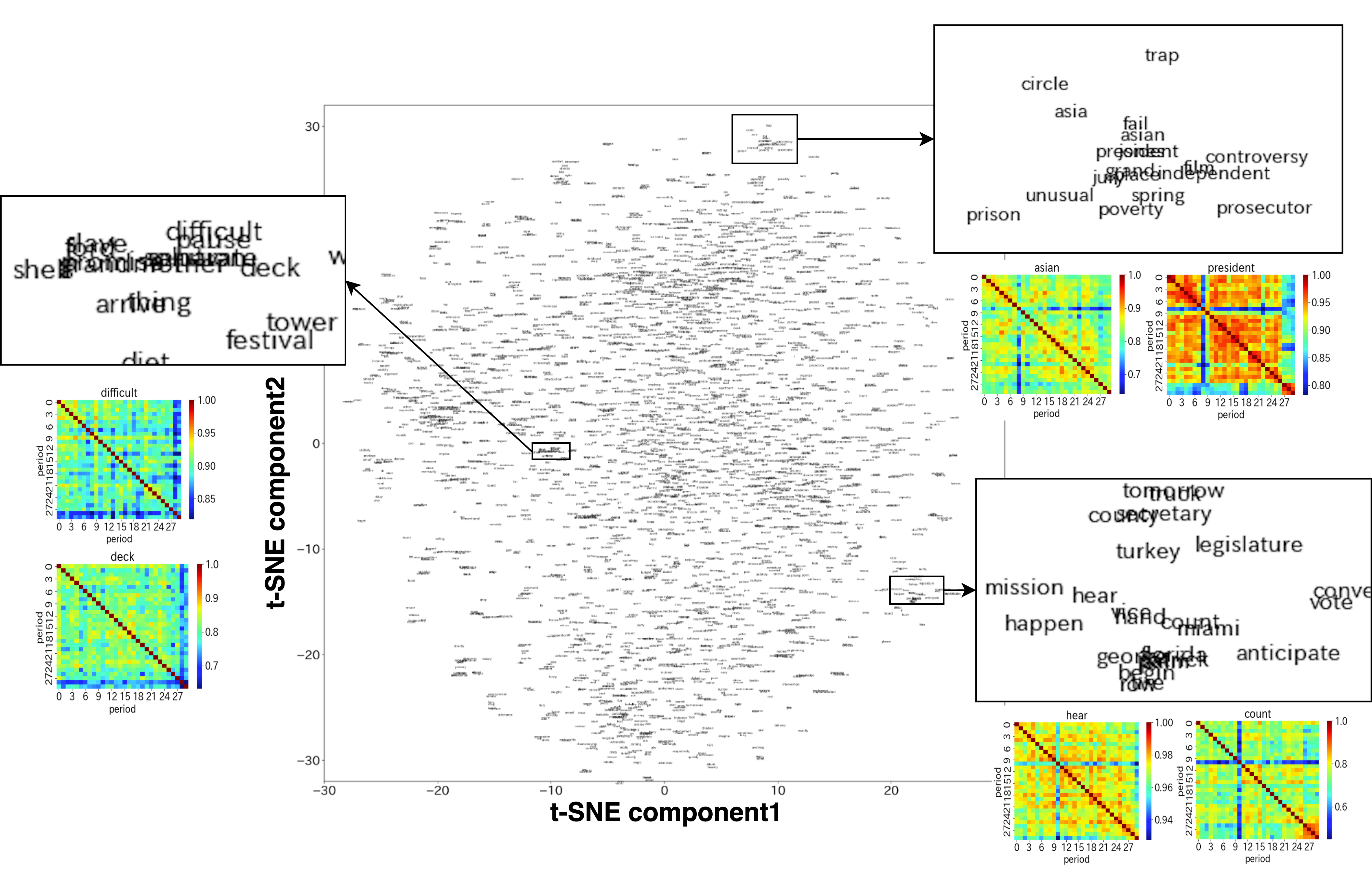}
    \caption{Visualizing the similarity matrix of all words in COCA using t-SNE in two dimensions shows that words close to each other in the compressed dimensions exhibit similar similarity patterns.  }
    \label{fig:tsne-coca}
\end{figure*}

\autoref{fig:tsne-coca} shows the result of visualizing the similarity matrix of all words in COCA in two dimensions using t-SNE. 
As with COHA, the similarity matrix compressed by t-SNE reveals that words located at nearby coordinates exhibit similar similarity matrix patterns. 
In \autoref{fig:tsne-coca}, many localized clusters were observed, such as the word ``persident'', which shows spikes during specific periods.

\section{Real Data in Japanese}
\label{sec:appendix:japan}
The Mainichi Shimbun is a Japanese newspaper corpus, segmented into 1-year periods from 2003 to 2020, resulting in embeddings for 18 time periods. 
Japanese texts are tokenized by  MeCab with UniDic.
The numbers of target words in Mainichi Shimbun is 7,228.
The statistics for Mainichi Shimbun is shown in \autoref{tab:app-mai}.
We used a subsampling threshold of $1e^{-4}$.

\begin{figure}[t]
    \centering
    \includegraphics[keepaspectratio, scale=0.055]{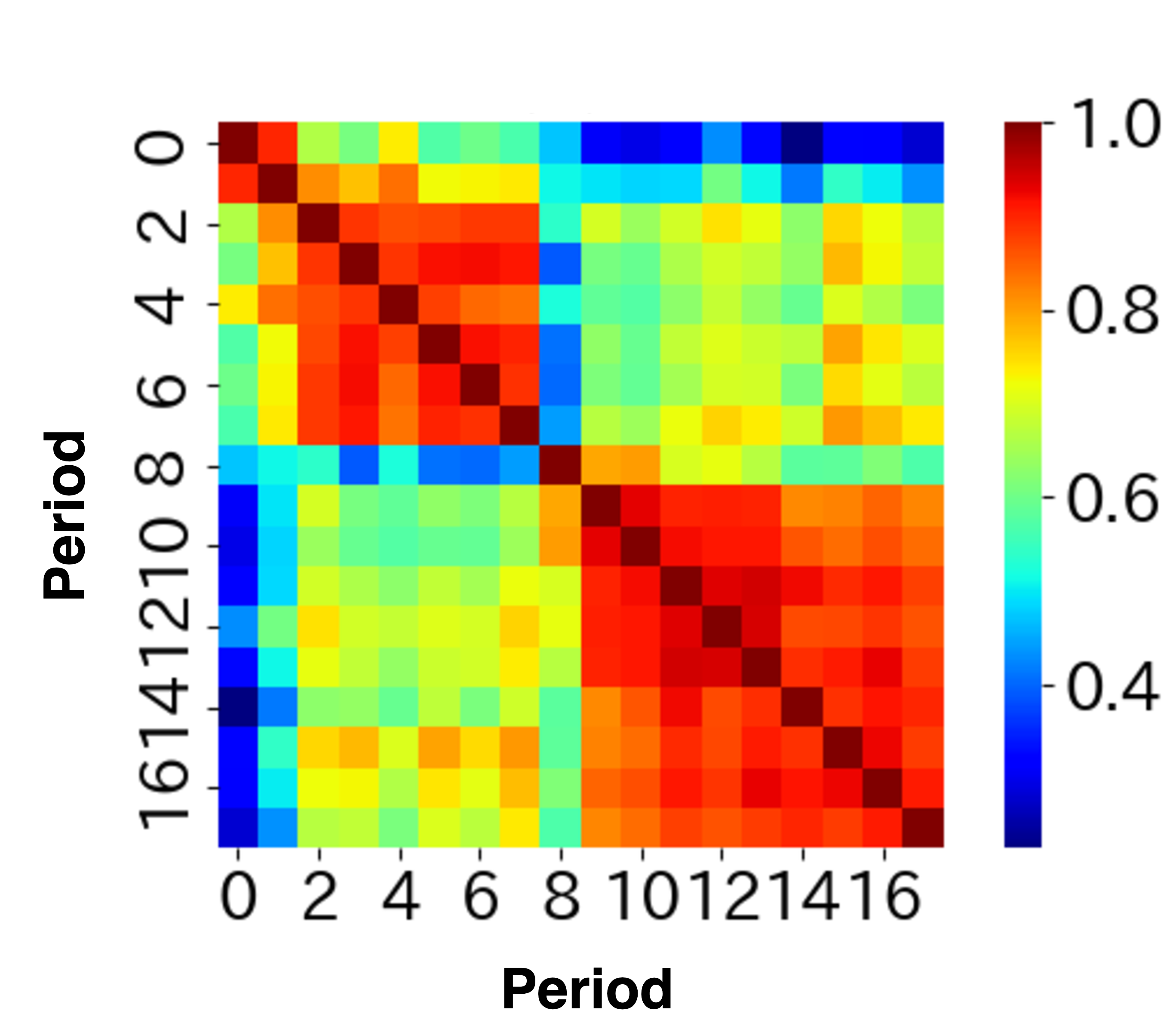}
    \caption{The similarity matrix of the word embeddings for the word ``復興'' (revival),learned from Mainichi Shimbun by PPMI-SVD joint. }
    \label{fig:cos_mai}
\end{figure}

\begin{table*}[t]
\small
\centering
\begin{tabular}{lllc} 
t1(target year) & t1  & co-occurrence word  & sense meaning   \\ \hline
2003 & 2005
& \begin{tabular}[c]{@{}l@{}l@{}l@{}l@{}}戦後 (postwar), 統治 (governance),\\ 主権 (sovereignty), 暫定 (provisional),\\ 協調 (cooperation), 主導 (leadership),\\ 仏 (France), 要員 (personnel),\\ 樹立 (establishment), 対外 (external/foreign)\end{tabular}              & Revival in Diplomatic Contexts       \\ \hline
\rule{0pt}{1.5em}%
2007 & 2003
& \begin{tabular}[c]{@{}l@{}l@{}l@{}l@{}}瓦礫 (rubble), 鐘 (bell),\\ 花火 (fireworks), 噴火 (eruption),\\ 津波 (tsunami), 地震 (earthquake),\\ 家屋 (house/building), 入居 (move-in),\\ 程遠い (far from), 古里 (hometown)\end{tabular}             & \begin{tabular}[c]{@{}l@{}}Revival in Response to Earthquakes and\\ Eruptions in Japan\end{tabular}       \\ \hline
\rule{0pt}{1.5em}%
2011 & 2007
& \begin{tabular}[c]{@{}l@{}l@{}l@{}l@{}l@{}}ビジョン (vision), 増税 (tax increase),\\ 提言 (proposal), 一元 (integration),\\ 歳出 (expenditure), 税 (tax), \\与野党 (ruling and opposition parties), \\県連 (prefectural federation),\\ 構想 (concept/plan), 税制 (tax system)\end{tabular} & \begin{tabular}[c]{@{}l@{}}Revival in Policy Discussions During \\the Great East Japan Earthquake\end{tabular} \\ \hline
\rule{0pt}{1.5em}%
2015 & 2011
& \begin{tabular}[c]{@{}l@{}l@{}l@{}l@{}}退去 (evacuation/departure), 資材 (materials),\\ 入居 (move-in), 全額 (full amount),\\ 開通 (opening to traffic),  井戸 (well),\\ 人道 (humanitarian), 帰還 (return),\\ 高騰 (surge/rise), 仮設 (temporary)\end{tabular}             & \begin{tabular}[c]{@{}l@{}}Revival Related to Policies Implemented \\After the Great East Japan Earthquake\end{tabular} \\ \hline    
\end{tabular}
\caption{A table showing the top 10 words and their meanings sorted by the differences in PPMI($\mathcal{N}_{10}^{(t_2 \rightarrow t_1)}$) for the word ``復興'' (revival) across four time periods: 2003, 2007, 2011, and 2015, learned from the Mainichi Shimbun dataset. 
}
\label{tab:hukkou}
\end{table*}

\begin{figure}[t]
    \centering
    \includegraphics[keepaspectratio, scale=0.4]{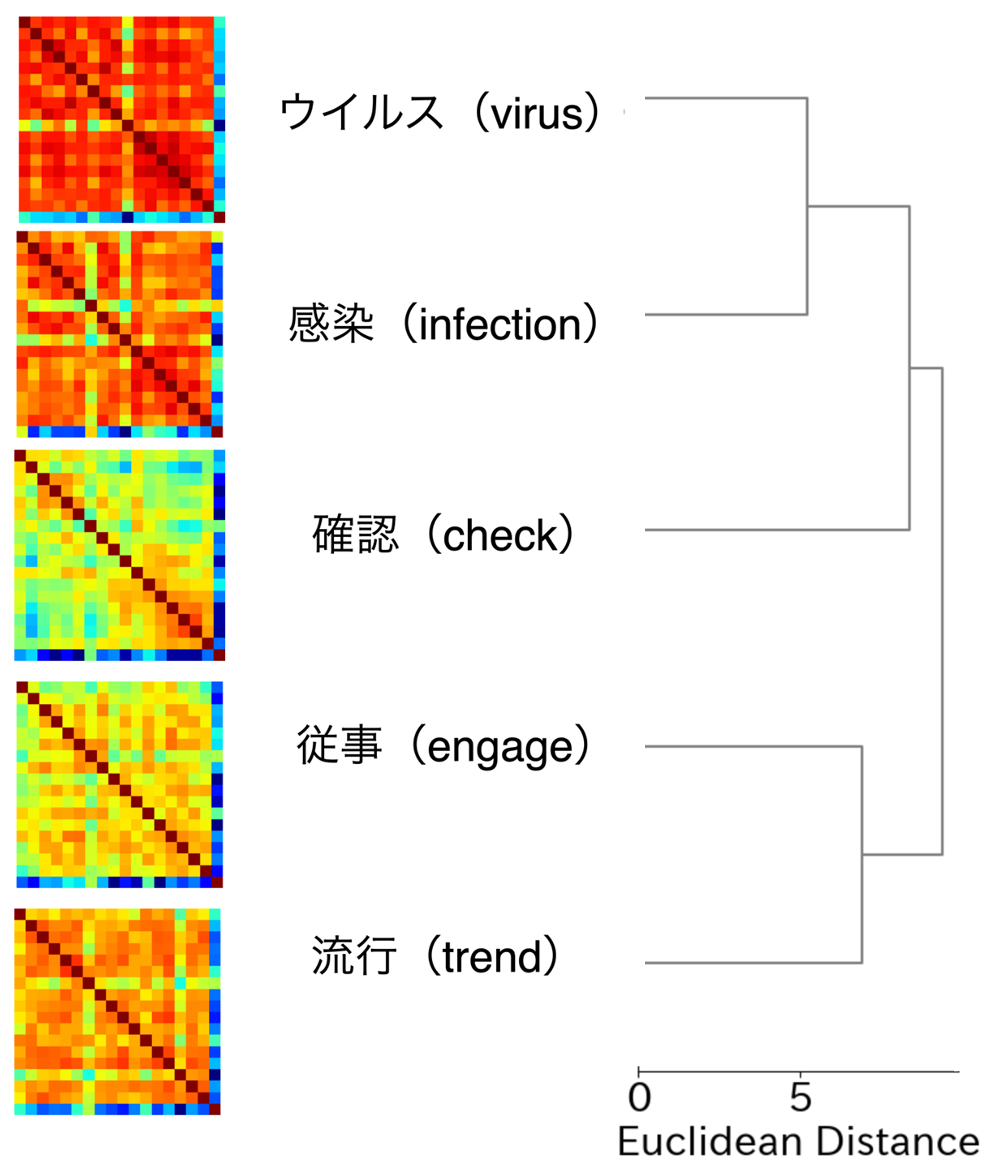}
    \caption{The results of the hierarchical clustering for words that are included in the same cluster as the word ``流行''(trend) in Mainichi Shimbun.}
    \label{fig:mai-cluster}
\end{figure}

\autoref{fig:cos_mai} presents the similarity matrix for the word ``復興'' (revival) learned from the Mainichi Shimbun. 
The word ``復興'' has not shifted in meaning but is believed to have experienced significant contextual shifts following the 2011 Great East Japan Earthquake, which is why it was selected.
It is evident that there are three clusters of high similarity, and the similarity between period 8 (2011) and other periods is low. This suggests that there are four distinct usages of the word ``復興''.
Compared to COHA and COCA, where the analysis is conducted in English, it is evident that similarity matrix analysis is also feasible in the case of Japanese.

\autoref{tab:hukkou} presents the top 10 words with the greatest differences in PPMI for the word ``復興'' (revival) in the Mainichi Shimbun dataset for the years 2003, 2007, 2011, and 2015.
In 2003, the word ``復興'' is associated with diplomatic contexts, as evidenced by co-occurring words such as ``主権'' (sovereignty), ``統治'' (governance), and ``仏'' (France).
In 2007, ``復興'' pertains to domestic natural disasters, with co-occurring words like ``噴火'' (eruption), ``地震'' (earthquake), and ``瓦礫'' (rubble), reflecting its use in the context of recovery from such events in Japan.
In 2011, following the Great East Japan Earthquake, ``復興'' is associated with political discussions, with co-occurring words such as ``ビジョン'' (vision), ``増税'' (tax increase), and ``構想'' (plan).
In 2015, ``復興'' is used in relation to the results of post-disaster policies, with co-occurring words like ``退去'' (evacuation), ``資材'' (materials), and ``開通'' (opening), reflecting the ongoing recovery efforts and their outcomes.

In \autoref{fig:mai-cluster}, the results of the hierarchical clustering for words included in the same cluster as the word ``流行'' (fashion/trend) are shown. It is observed that ``流行'' is clustered with words like ``感染'' (infection) , ``確認'' (check), ``ウイルス'' (virus) and ``従事'' (engage). These words are relevant to ``流行'', indicating that the clustering effectively categorizes trends in semantic shifts. This cluster can be interpreted as one where the context of usage has shifted due to the impact of COVID-19.

From the experimental results mentioned above, it is evident that this framework is also useful for analyzing semantic shifts in the Japanese language.

\section{Pseudo Data}
\subsection{Detail Setup}
\label{sec:appendix:pseudo-setting}

The probability for each sense is calculated as described below. In C1, $sense_1$ remains constant at 0.7, while $sense_2$ increases from 0.1 to 1 on a logarithmic scale. In C2, $sense_1$ decreases from 1 to 0.1 on a logarithmic scale, and $sense_2$ increases from 0.1 to 1 on a logarithmic scale. In C3, $sense_1$ increases from 0.1 to 1 on a logarithmic scale, and the $senses$ are sampled from a Dirichlet distribution for each of the seven meanings. In D1, $sense_1$ decreases from 1 to 0.1 on a logarithmic scale. In D2, $sense_1$ remains constant at 0.7, and $sense_2$ shows spikes at 0.55 during periods 4 and 6, with a value of 0.1 otherwise. In D3, $sense_1$ remains constant at 0.7, and $sense_2$ exhibits periodic spikes during periods 1 and 3, 7 and 9, and 13 and 15, with a value of 0.1 otherwise. In D4, each of the $senses$ is sampled from a Dirichlet distribution.The implementation of the logarithmic scale and Dirichlet distribution was carried out using Numpy.

\begin{figure*}[t]
    \begin{tabular}{c}
    \centering    
      \begin{minipage}[t]{0.33\hsize}
        \centering
        \includegraphics[keepaspectratio, scale=0.3]{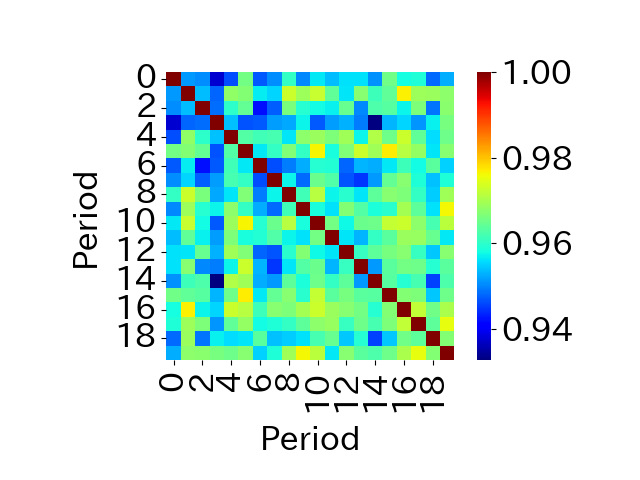}
        \subcaption{pseudoword "C1-0-4"}
      \end{minipage} 
      \begin{minipage}[t]{0.33\hsize}
        \centering
        \includegraphics[keepaspectratio, scale=0.3]{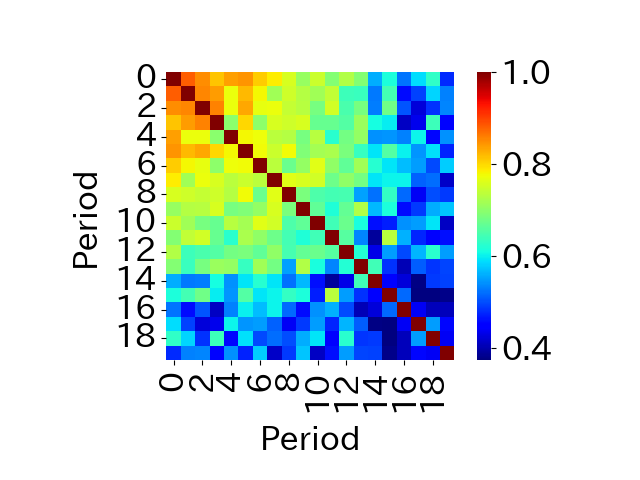}
        \subcaption{pseudoword "C2-1-3"}
      \end{minipage} 
      \begin{minipage}[t]{0.33\hsize}
        \centering
        \includegraphics[keepaspectratio, scale=0.3]{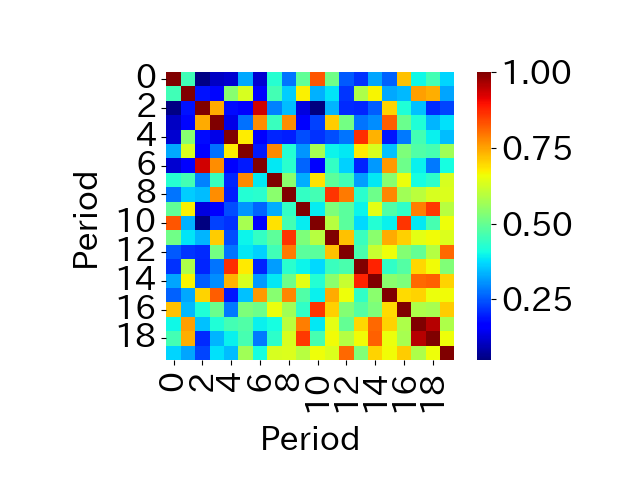}
        \subcaption{pseudoword "C3-1-0"}
      \end{minipage} 
    \end{tabular}
    \begin{tabular}{cc}
      \centering
      \begin{minipage}[t]{0.24\hsize}
        \centering
        \includegraphics[keepaspectratio, scale=0.27]{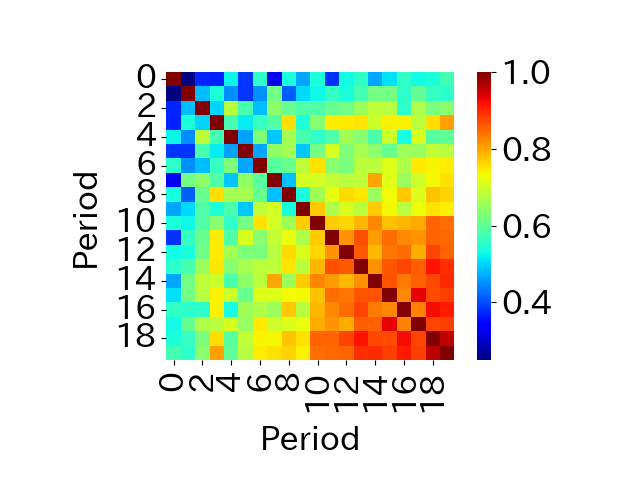}
        \subcaption{pseudoword "D1-2-4"}
      \end{minipage} 
      \begin{minipage}[t]{0.24\hsize}
        \centering
        \includegraphics[keepaspectratio, scale=0.27]{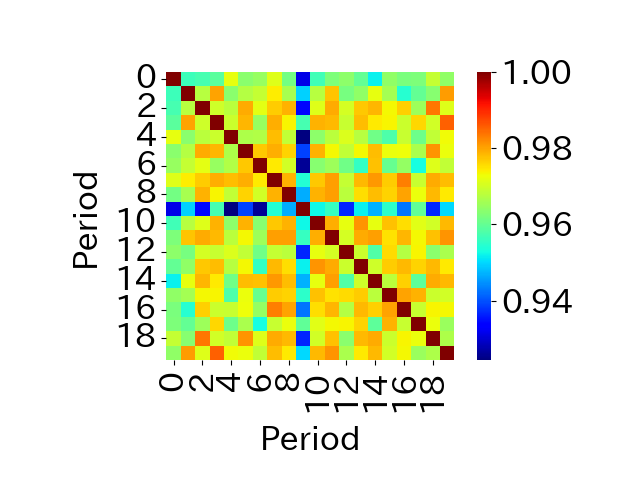}
        \subcaption{pseudoword "D2-1-1"}
      \end{minipage} 
      \begin{minipage}[t]{0.24\hsize}
        \centering
        \includegraphics[keepaspectratio, scale=0.27]{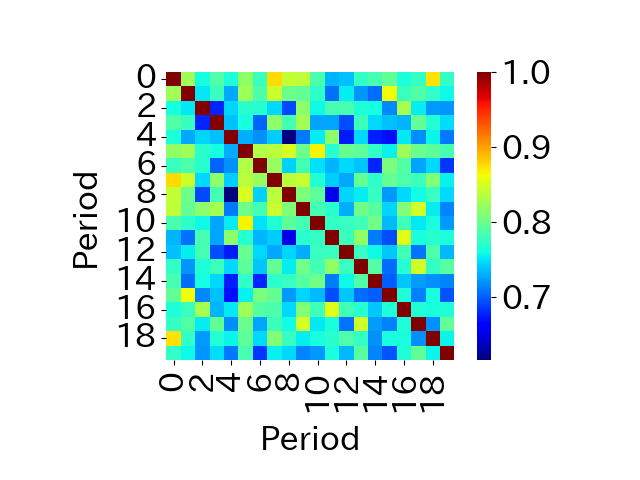}
        \subcaption{pseudoword "D3-2-0"}
      \end{minipage} 
      \begin{minipage}[t]{0.24\hsize}
        \centering
        \includegraphics[keepaspectratio, scale=0.27]{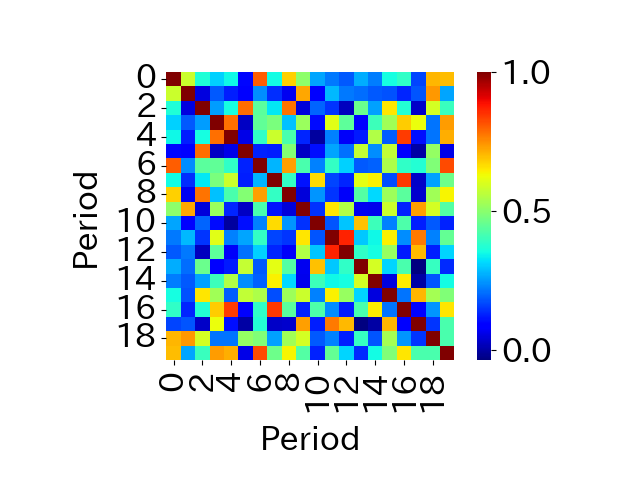}
        \subcaption{pseudoword "D4-2-0"}
      \end{minipage} 
    \end{tabular}
     \caption{The similarity matrices were visualized for each shift schema. The similarity matrices generally captured the characteristics of the schemas and could be interpreted manually.}
     \label{fig:pseuod-cos}
\end{figure*}

\begin{figure}[tb]
    \centering
    \includegraphics[scale=0.25]{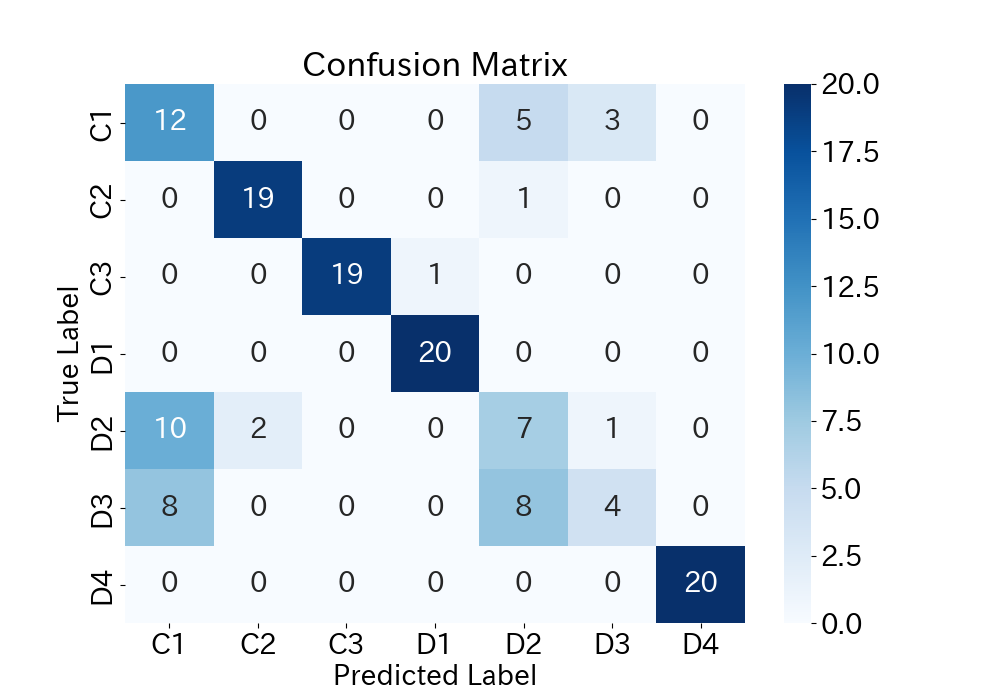}
    \caption{The confusion matrix resulting from hierarchical clustering using the standardized upper triangular matrix.}
    \label{fig:conf}
\end{figure}

\begin{figure*}[t]
    \begin{tabular}{cc}
      \centering
      \begin{minipage}[t]{0.24\hsize}
        \centering
        \includegraphics[keepaspectratio, scale=0.15]{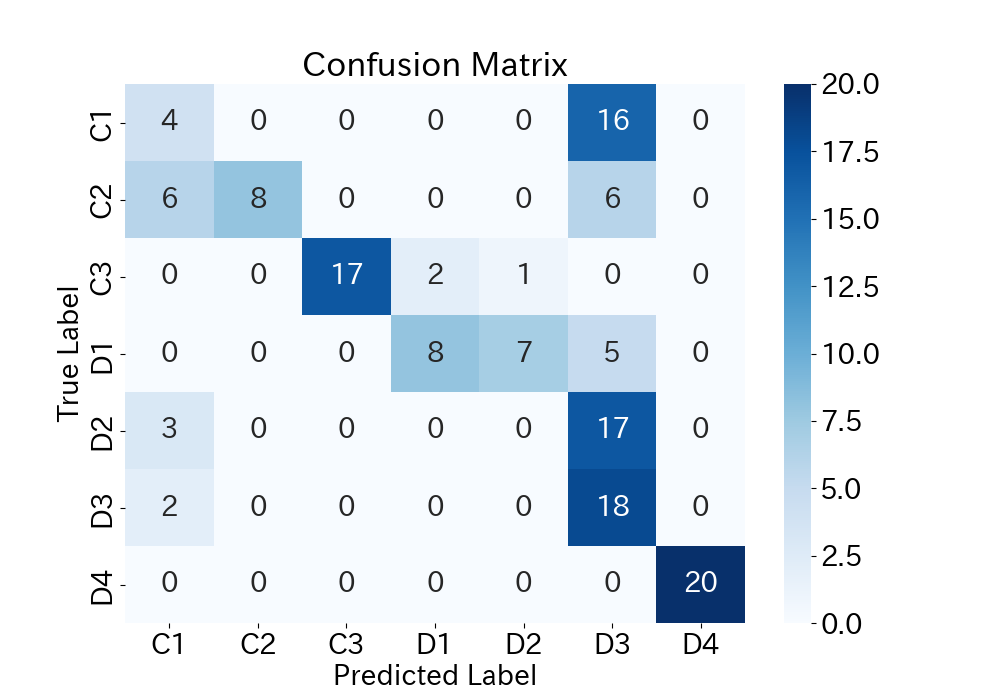}
        \subcaption{cos+agglo+adj}
      \end{minipage} 
      \begin{minipage}[t]{0.24\hsize}
        \centering
        \includegraphics[keepaspectratio, scale=0.15]{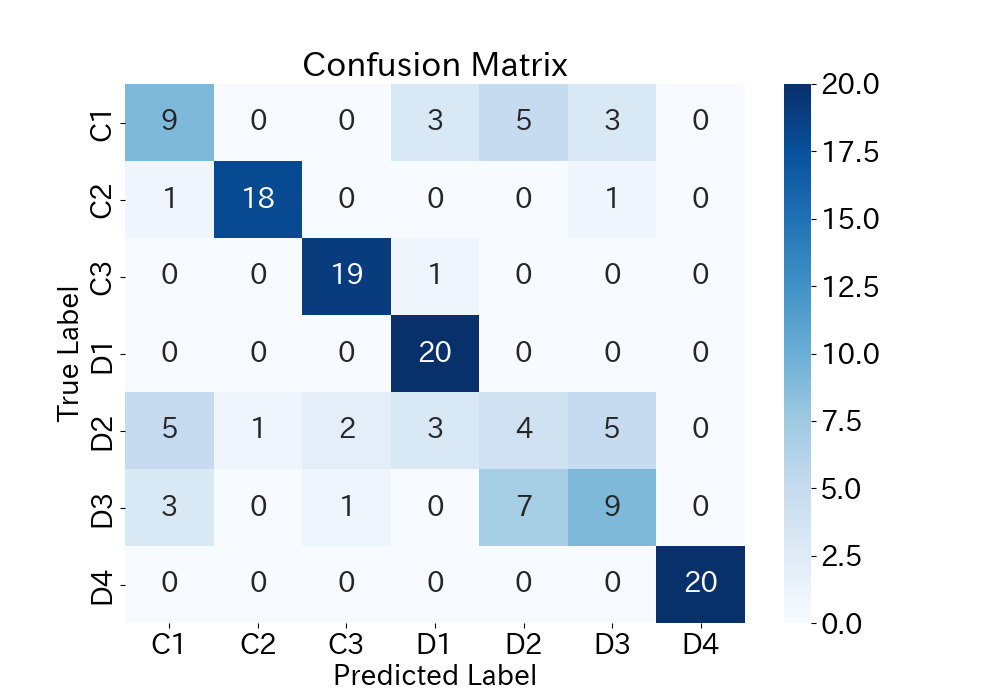}
        \subcaption{cos+agglo+adj+z}
      \end{minipage} 
      \begin{minipage}[t]{0.24\hsize}
        \centering
        \includegraphics[keepaspectratio, scale=0.15]{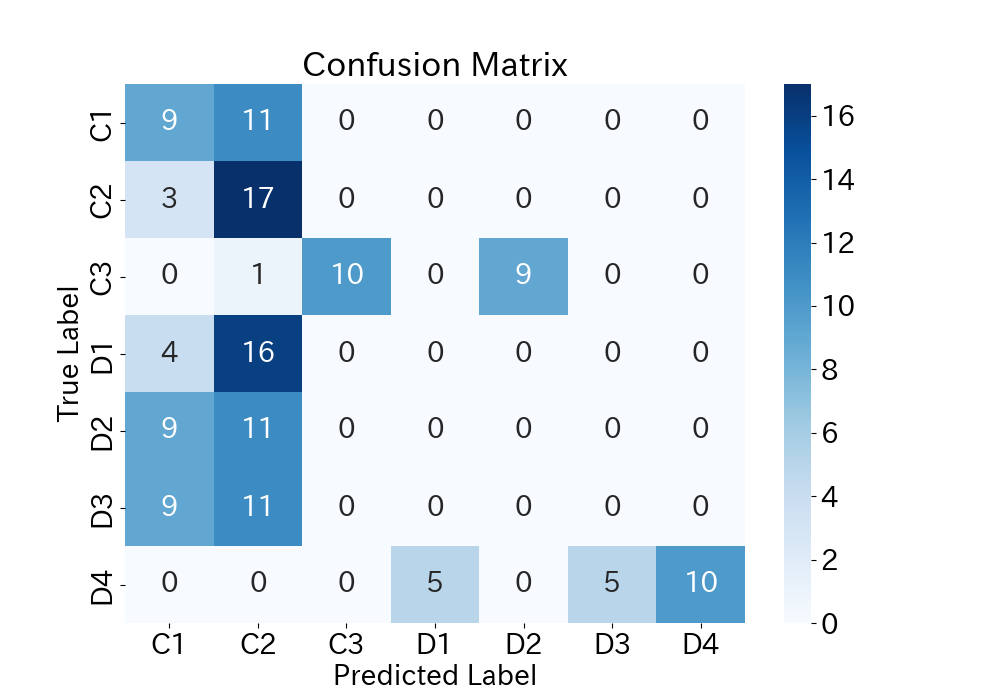}
        \subcaption{eu+agglo+adj}
      \end{minipage} 
      \begin{minipage}[t]{0.24\hsize}
        \centering
        \includegraphics[keepaspectratio, scale=0.15]{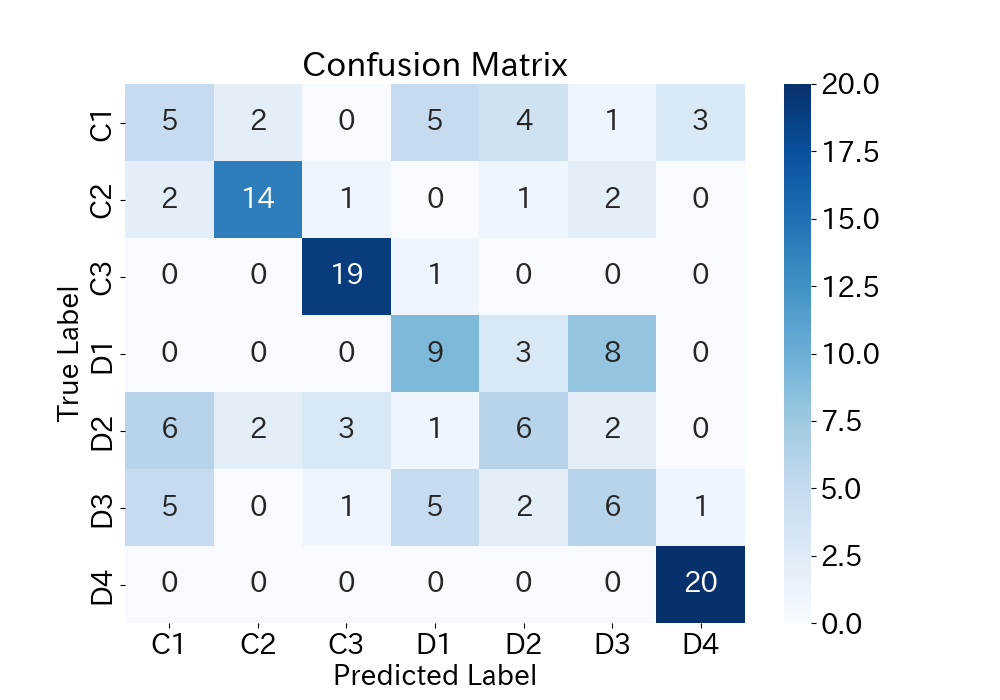}
        \subcaption{eu+agglo+adj+z}
      \end{minipage} 
    \end{tabular}
    \begin{tabular}{cc}
      \centering
      \begin{minipage}[t]{0.24\hsize}
        \centering
        \includegraphics[keepaspectratio, scale=0.15]{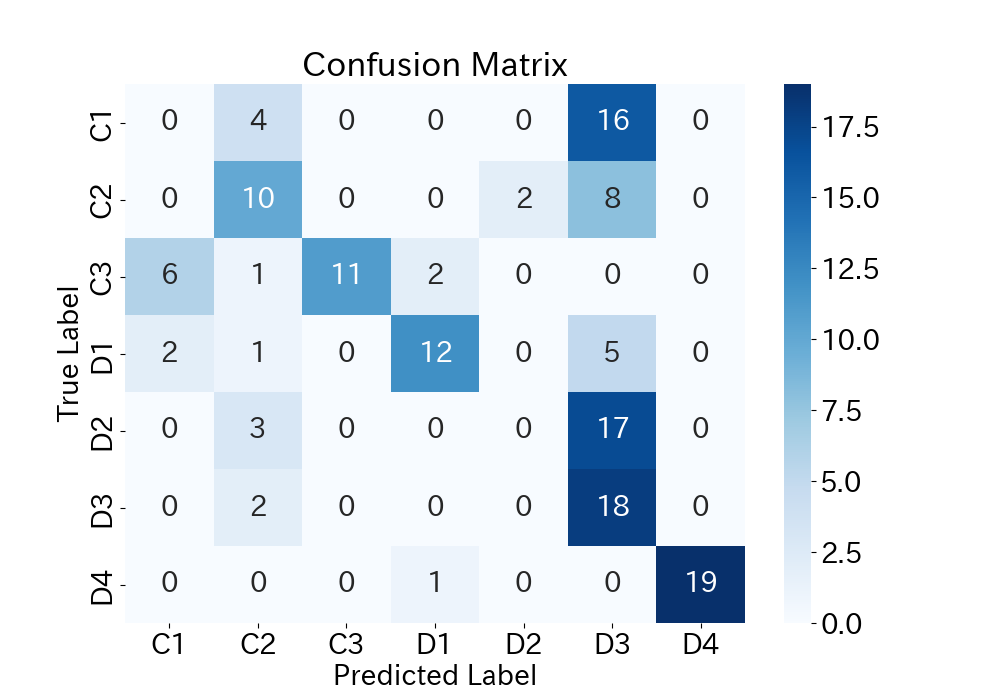}
        \subcaption{cos+agglo+time0}
      \end{minipage} 
      \begin{minipage}[t]{0.24\hsize}
        \centering
        \includegraphics[keepaspectratio, scale=0.15]{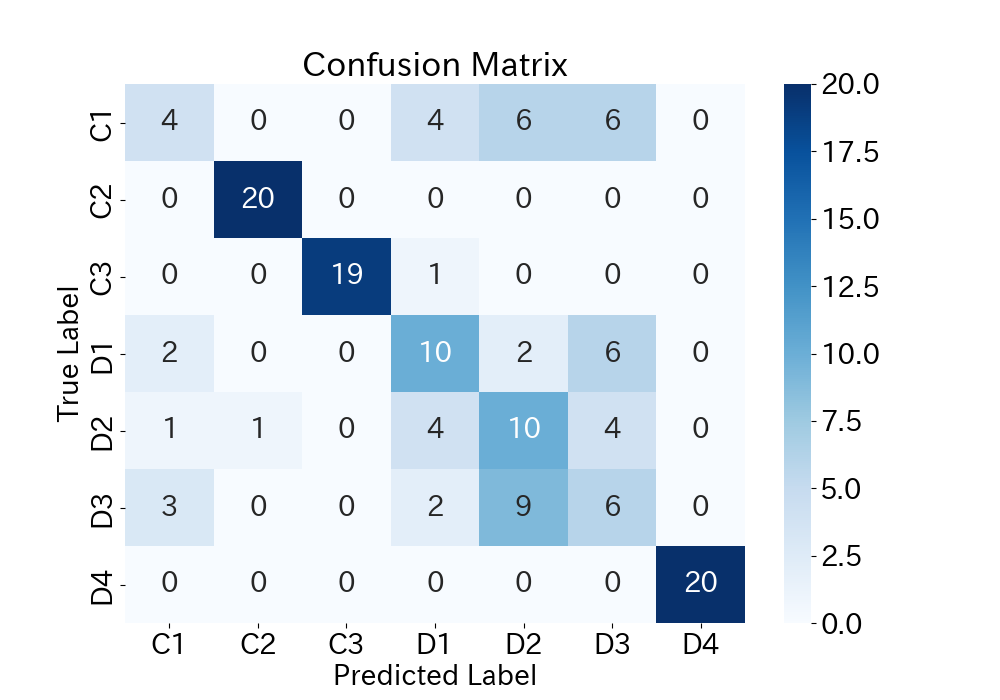}
        \subcaption{cos+agglo+time0+z}
      \end{minipage} 
      \begin{minipage}[t]{0.24\hsize}
        \centering
        \includegraphics[keepaspectratio, scale=0.15]{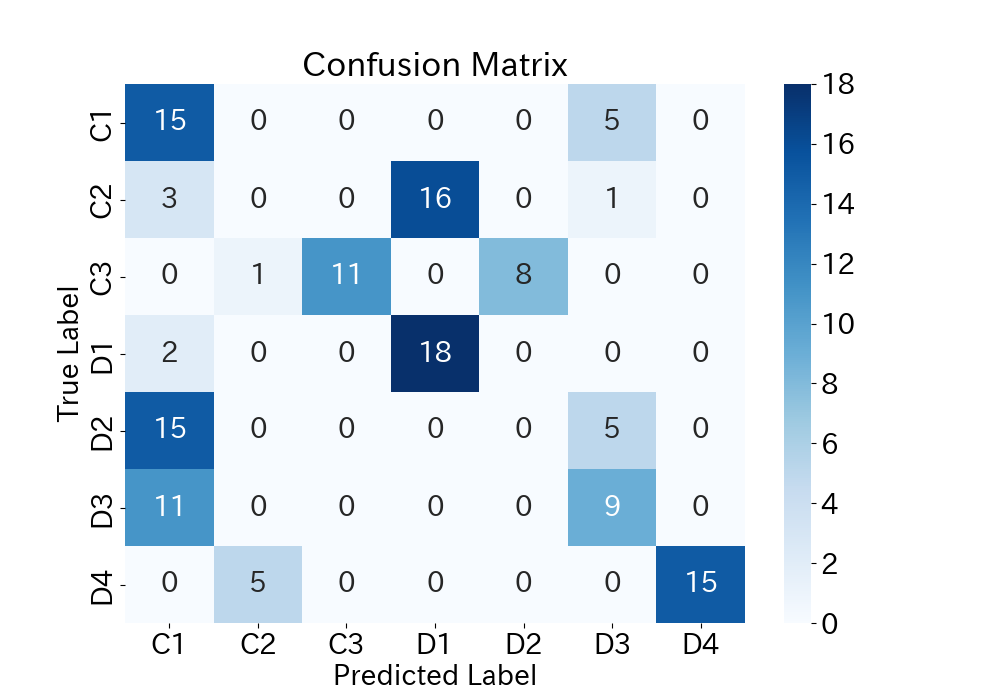}
        \subcaption{eu+agglo+time0}
      \end{minipage} 
      \begin{minipage}[t]{0.24\hsize}
        \centering
        \includegraphics[keepaspectratio, scale=0.15]{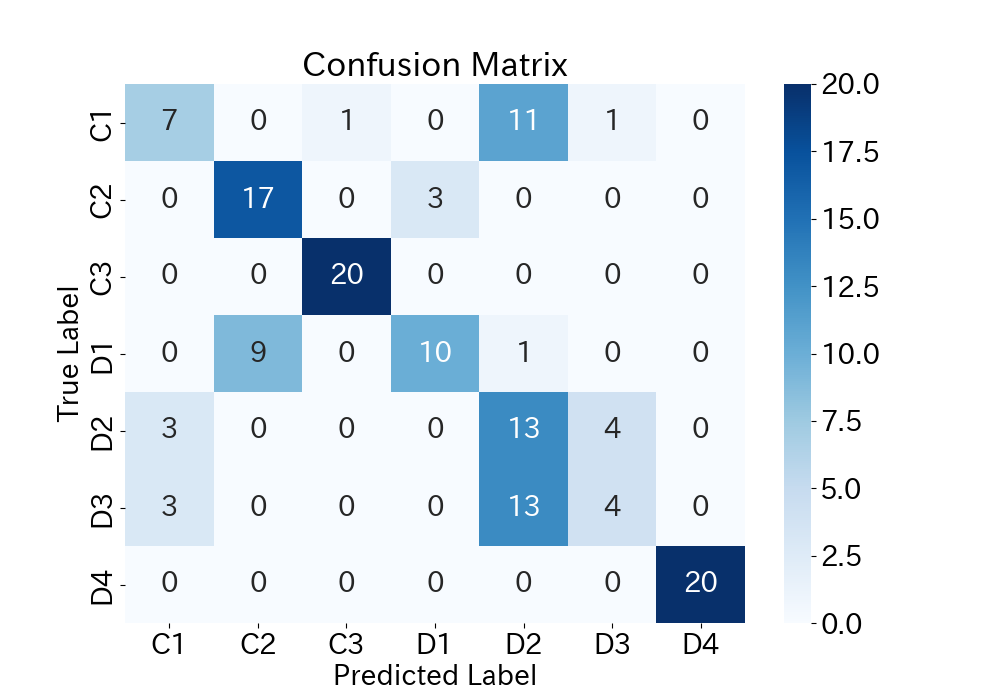}
        \subcaption{eu+agglo+time0+z}
      \end{minipage} 
    \end{tabular}
    \begin{tabular}{cc}
      \centering
      \begin{minipage}[t]{0.24\hsize}
        \centering
        \includegraphics[keepaspectratio, scale=0.15]{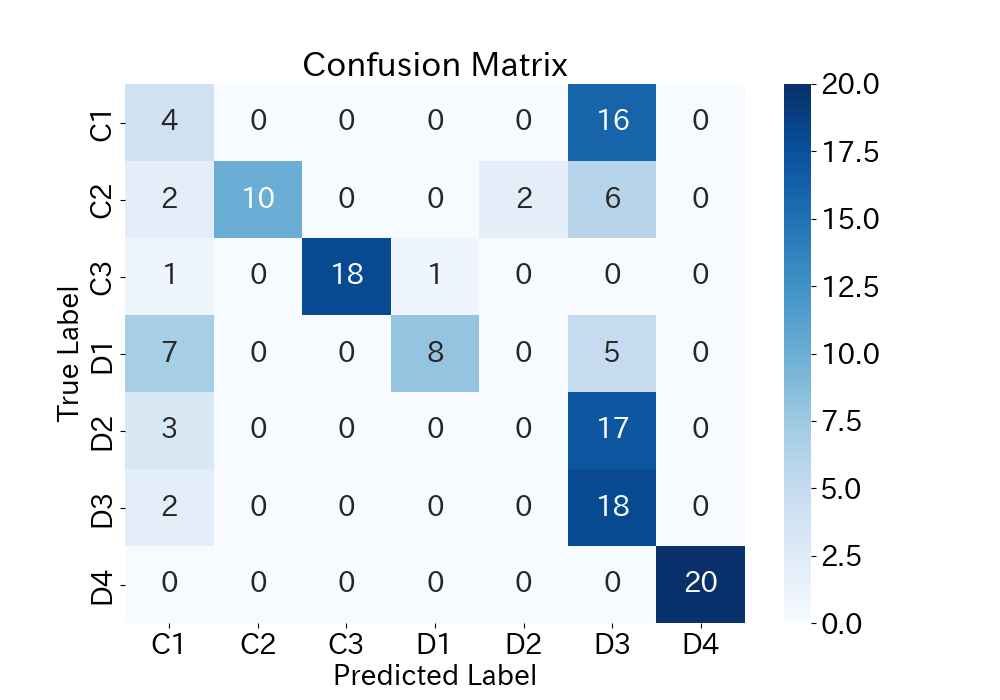}
        \subcaption{cos+agglo+tri}
      \end{minipage} 
      \begin{minipage}[t]{0.24\hsize}
        \centering
        \includegraphics[keepaspectratio, scale=0.15]{fig/conf/agglo_tri_z.png}
        \subcaption{cos+agglo+tri+z}
      \end{minipage} 
      \begin{minipage}[t]{0.24\hsize}
        \centering
        \includegraphics[keepaspectratio, scale=0.15]{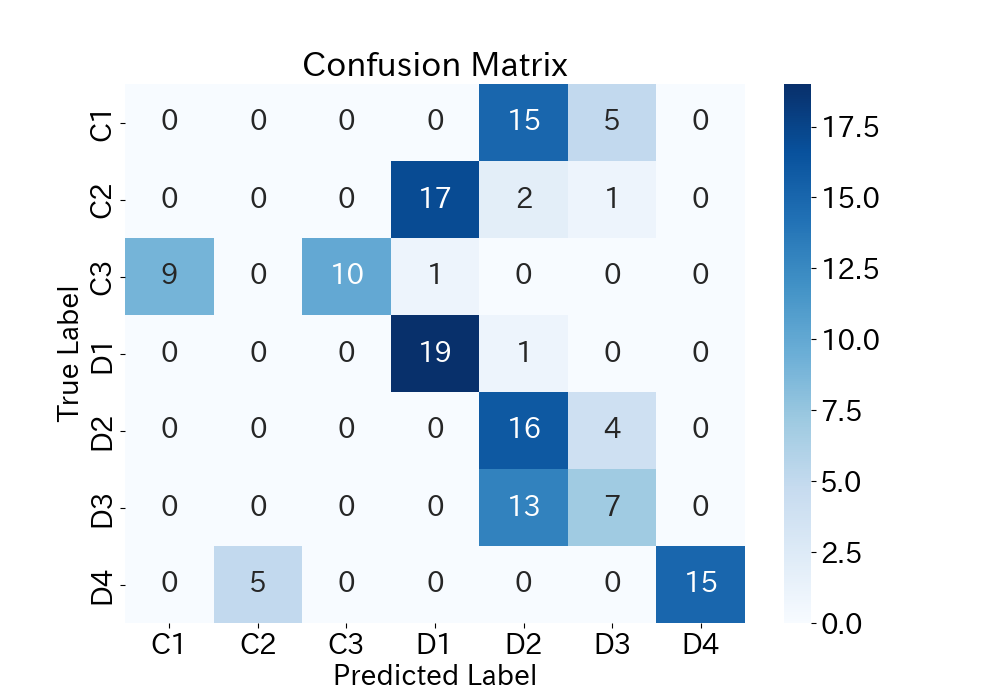}
        \subcaption{eu+agglo+tri}
      \end{minipage} 
      \begin{minipage}[t]{0.24\hsize}
        \centering
        \includegraphics[keepaspectratio, scale=0.15]{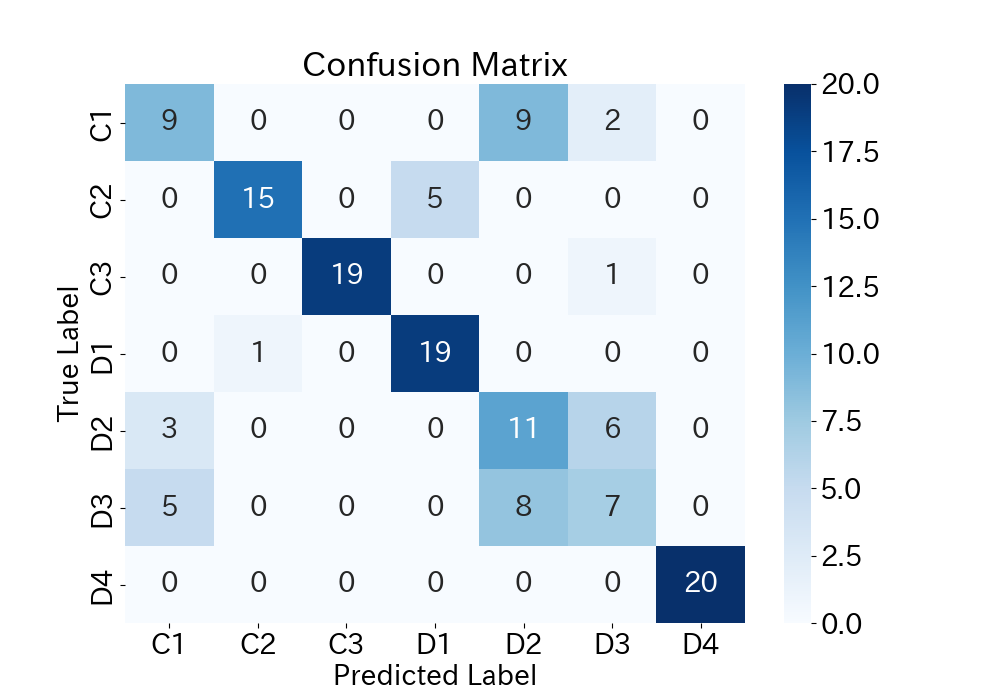}
        \subcaption{eu+agglo+tri+z}
      \end{minipage} 
    \end{tabular}
    \begin{tabular}{cc}
      \centering
      \begin{minipage}[t]{0.24\hsize}
        \centering
        \includegraphics[keepaspectratio, scale=0.15]{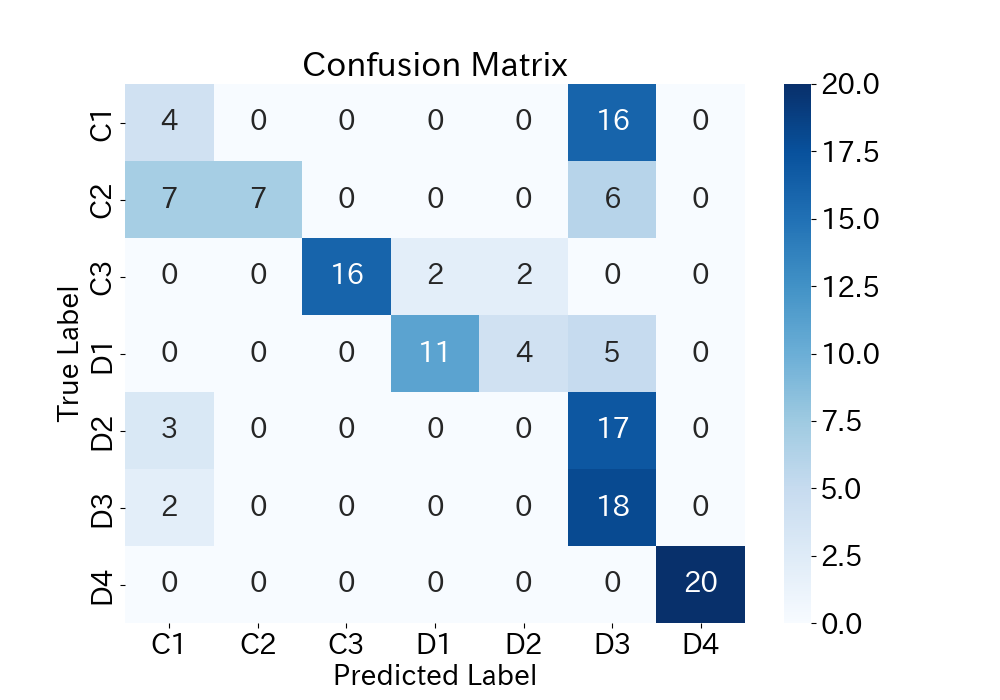}
        \subcaption{cos+kmeans+adj}
      \end{minipage} 
      \begin{minipage}[t]{0.24\hsize}
        \centering
        \includegraphics[keepaspectratio, scale=0.15]{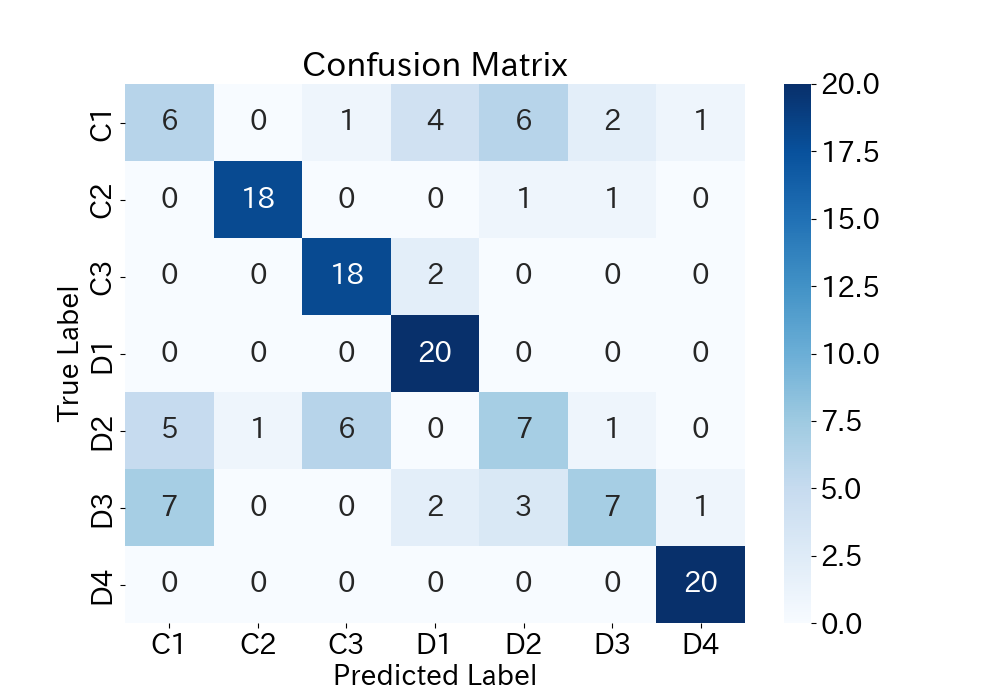}
        \subcaption{cos+kmeans+adj+z}
      \end{minipage} 
      \begin{minipage}[t]{0.24\hsize}
        \centering
        \includegraphics[keepaspectratio, scale=0.15]{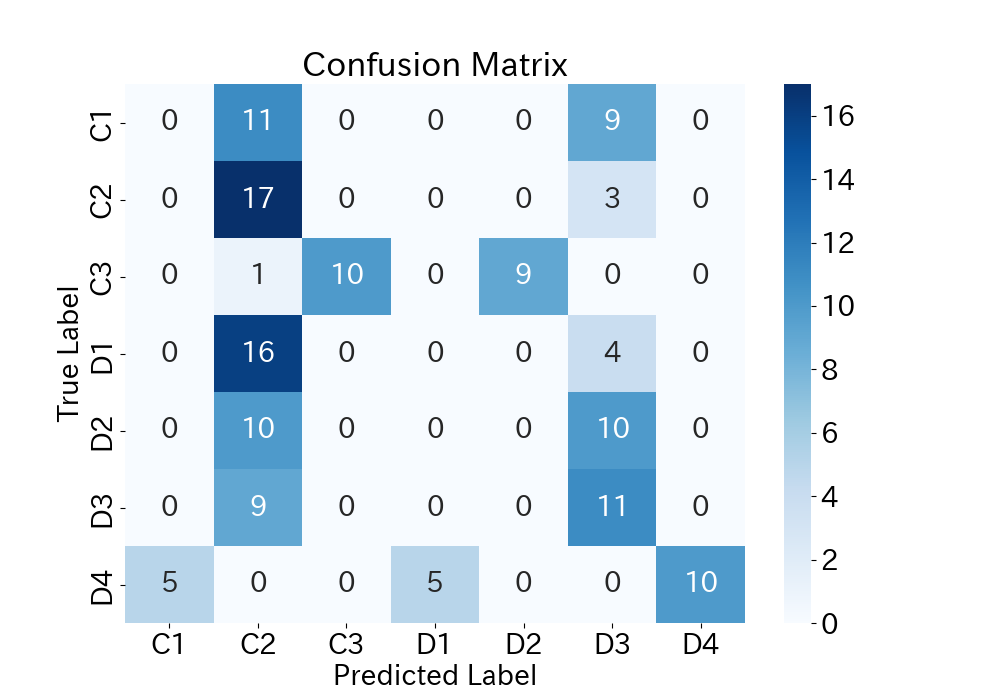}
        \subcaption{eu+adj+kmeans}
      \end{minipage} 
      \begin{minipage}[t]{0.24\hsize}
        \centering
        \includegraphics[keepaspectratio, scale=0.15]{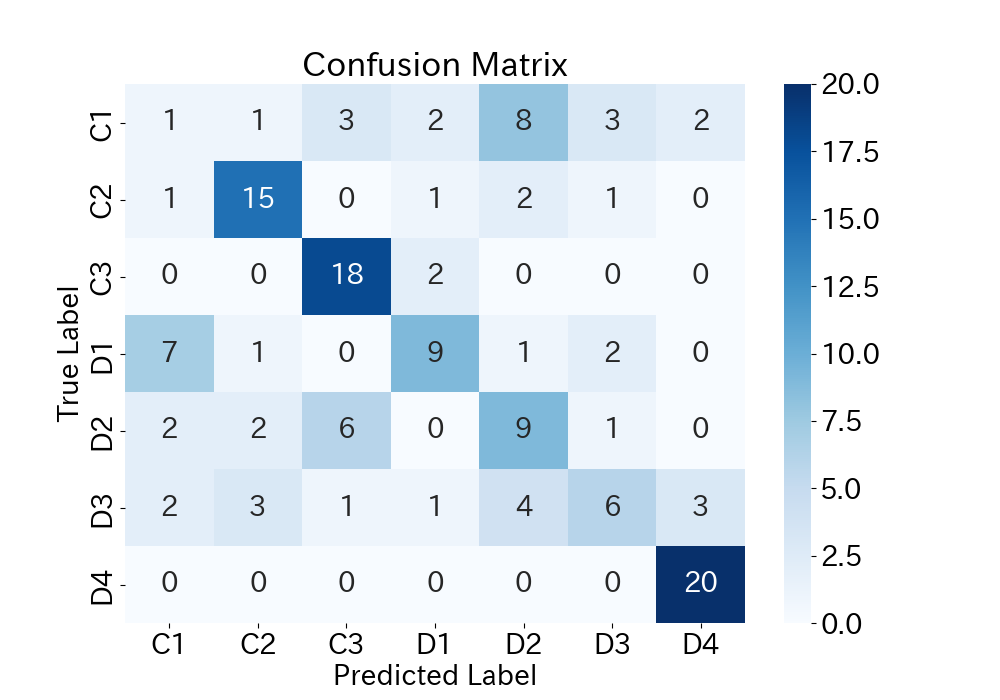}
        \subcaption{eu+kmeans+adj+z}
      \end{minipage} 
    \end{tabular}
    \begin{tabular}{cc}
      \centering
      \begin{minipage}[t]{0.24\hsize}
        \centering
        \includegraphics[keepaspectratio, scale=0.15]{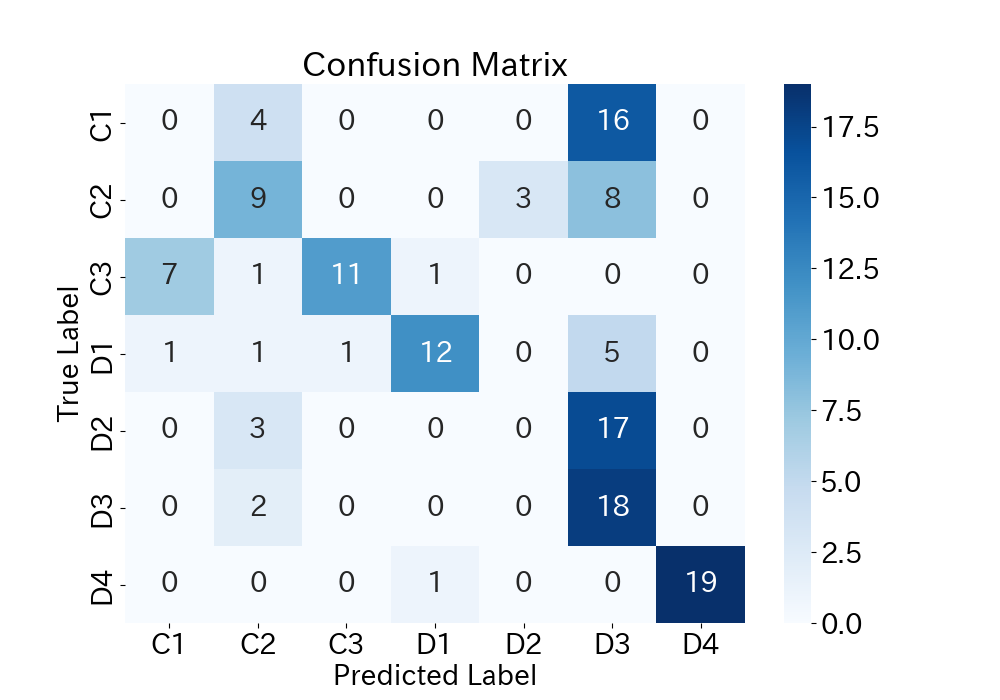}
        \subcaption{cos+kmeans+time0}
      \end{minipage} 
      \begin{minipage}[t]{0.24\hsize}
        \centering
        \includegraphics[keepaspectratio, scale=0.15]{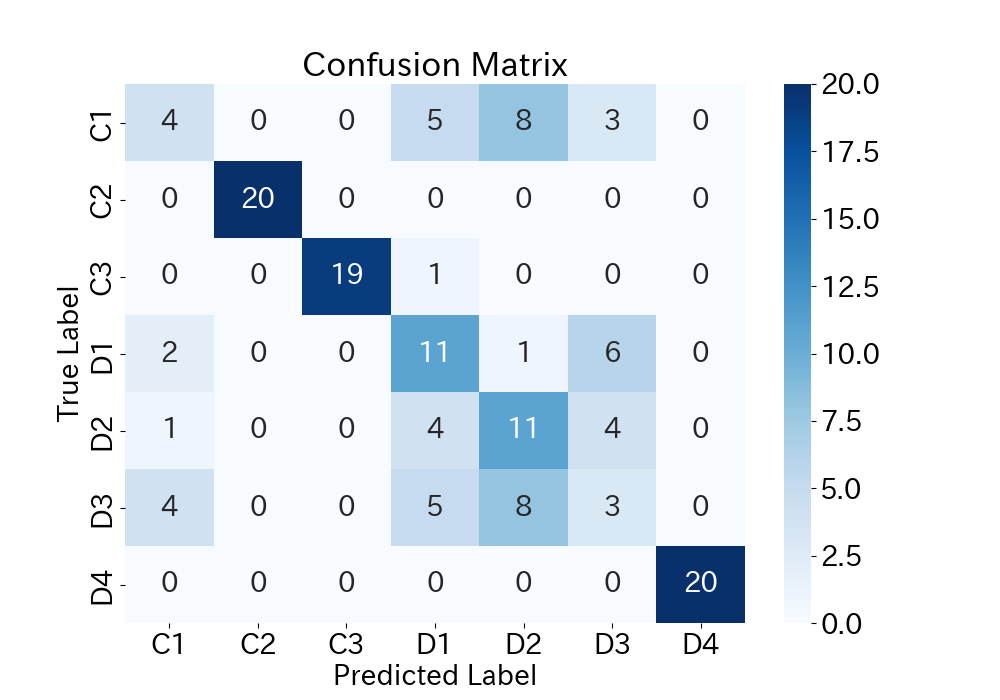}
        \subcaption{cos+kmeans+time0+z}
      \end{minipage} 
      \begin{minipage}[t]{0.24\hsize}
        \centering
        \includegraphics[keepaspectratio, scale=0.15]{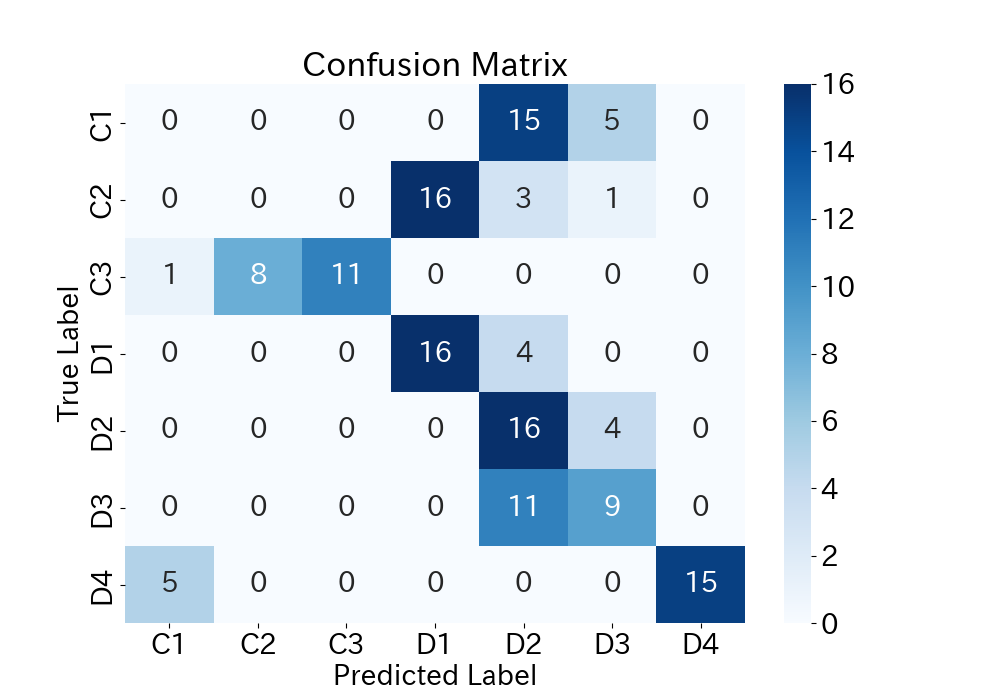}
        \subcaption{eu+kmeans+time0}
      \end{minipage} 
      \begin{minipage}[t]{0.24\hsize}
        \centering
        \includegraphics[keepaspectratio, scale=0.15]{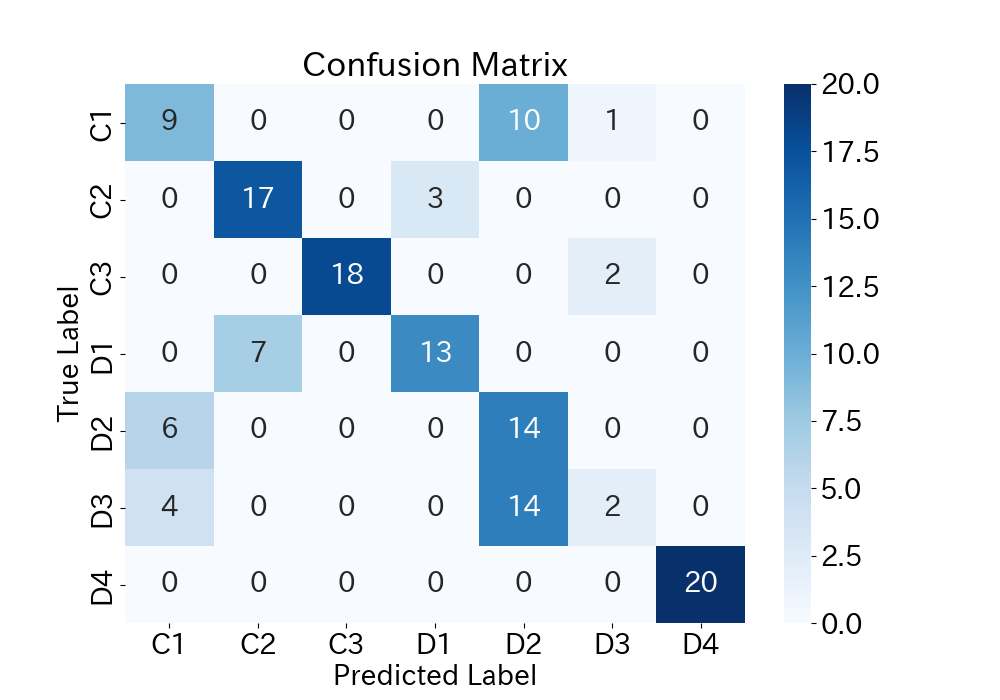}
        \subcaption{eu+kmeans+time0+z}
      \end{minipage} 
    \end{tabular}
    \begin{tabular}{cc}
      \centering
      \begin{minipage}[t]{0.24\hsize}
        \centering
        \includegraphics[keepaspectratio, scale=0.15]{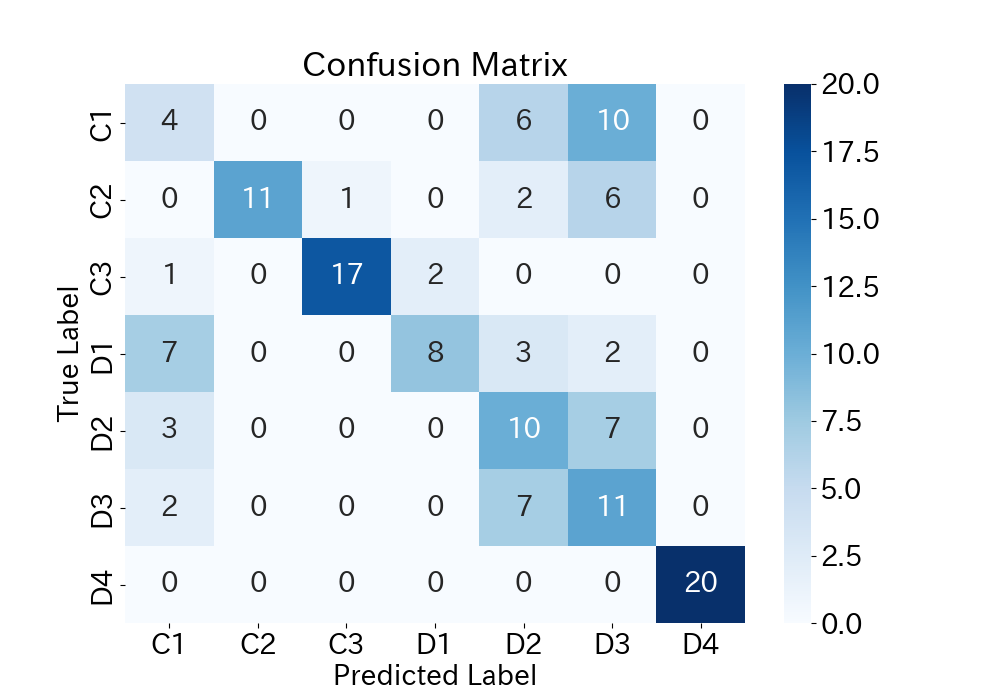}
        \subcaption{cos+kmeans+tri}
      \end{minipage} 
      \begin{minipage}[t]{0.24\hsize}
        \centering
        \includegraphics[keepaspectratio, scale=0.15]{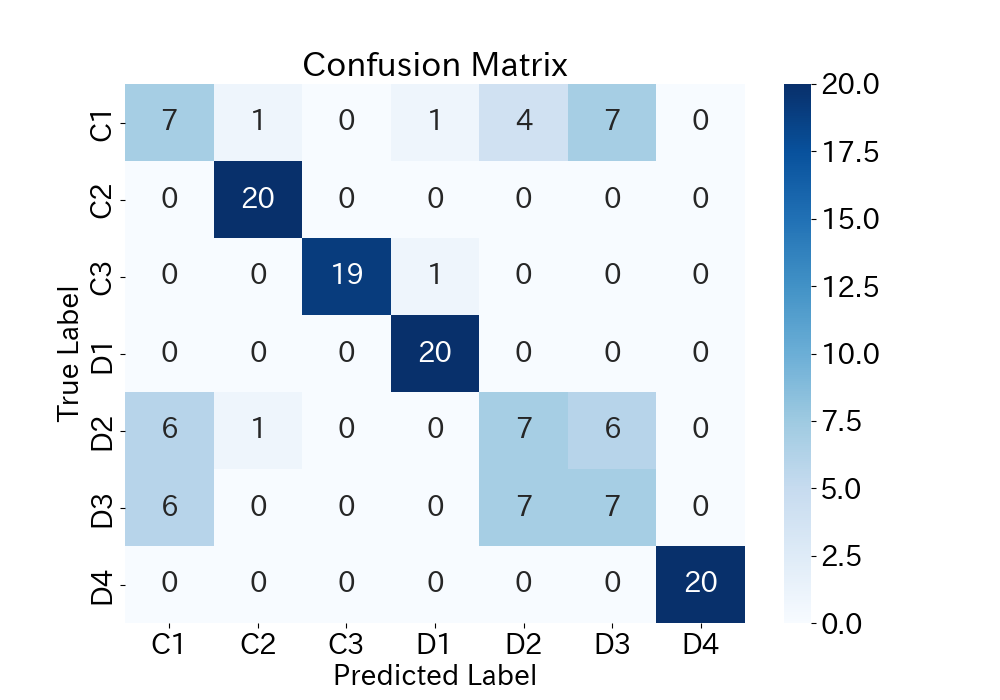}
        \subcaption{cos+kmeans+tri+z}
      \end{minipage} 
      \begin{minipage}[t]{0.24\hsize}
        \centering
        \includegraphics[keepaspectratio, scale=0.15]{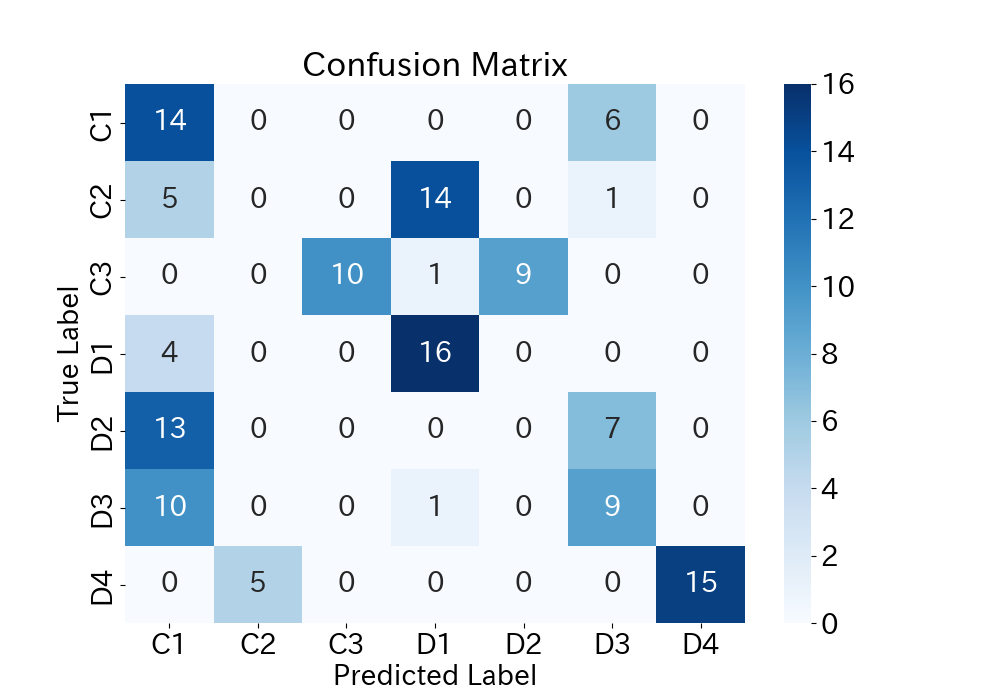}
        \subcaption{eu+kmeans+tri}
      \end{minipage} 
      \begin{minipage}[t]{0.24\hsize}
        \centering
        \includegraphics[keepaspectratio, scale=0.15]{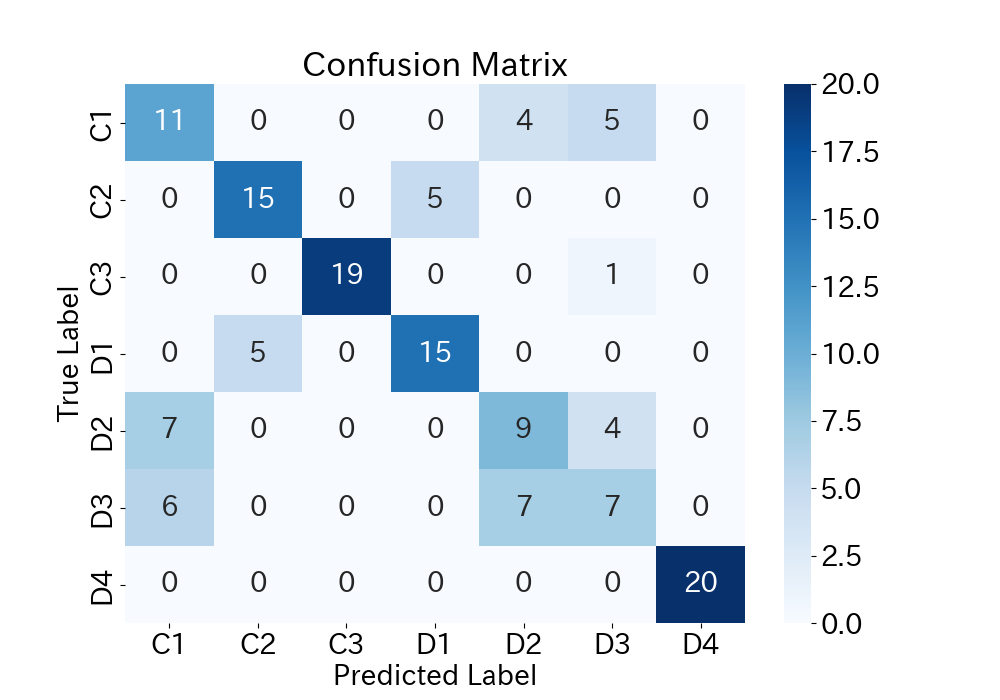}
        \subcaption{eu+kmeans+tri+z}
      \end{minipage} 
    \end{tabular}
     \caption{Confusion Matrices in classification of pseudo-shift schemas.}
     \label{fig:pseuod-all-conf}
\end{figure*}

\subsection{Analysis of Similarity Matrices}
\label{subsec:exp-pseudo_simmat}
We visualize and analyze the similarity matrices for each shift schema using pseudo data.
~\autoref{fig:pseuod-cos} shows the similarity matrices of word embeddings for pseudo-words corresponding to each shift schema. While some shifts are difficult to detect, such as in C1 and D3, others, like C2 and D1, reveal patterns where similarity increases or decreases, or where spikes occur at specific periods, as in D2. Shifts involving random noise, as seen in C3 and D4, can also be detected. 
This demonstrates the high interpretability of the similarity matrices.

However, it is important to note that frequency bias cannot be completely eliminated. For example, when comparing C2 and D1, if the time axis of C2 is reversed, its similarity matrix closely resembles that of D1. This indicates a tendency for periods with higher frequencies to exhibit greater similarity in embeddings.

\subsection{Confusion Matrices in Shift Schema}
\label{sec:appendix:confusion}

From the experimental results mentioned above, it was quantitatively demonstrated that the validity of the proposed method in classifying pseudo-shift schemas can be established by using and standardizing a large number of components in the similarity matrix.
We visualize the confusion matrix of the classification results and analyze which shift schemas can and cannot be identified.
\autoref{fig:conf} show the confusion matrices obtained by performing hierarchical clustering using the upper triangular matrices with standardization. 
From \autoref{fig:conf}, the schemas C2, C3, D1 and D4 tend to be well classified.
On the other hand, schemas such as C1, D2, and D3 could not be detected. 
The C1 schema involves the model of polysemous words acquiring new meanings. 
Because the proportion of $word_1$ remains constant, similarities increase, which may make detection difficult through similarity\-based analysis or indicate limitations of the embedding methods．
Schemas D2 and D3 involve spikes, and since the timing of these spikes is random, the shifts in similarity might not have clustered effectively.

We include the confusion matrices obtained from various methods in the analysis in \autoref{fig:pseuod-all-conf}.
We investigate the use of cosine similarity (cos) and Euclidean distance (eu) as methods for calculating similarity. 
The first two columns in the figure correspond to the use of cosine similarity, while the right two columns correspond to the use of Euclidean distance. Additionally, we examine the use of hierarchical clustering (agglo) and K-means (kmeans) as clustering methods. 
The top three rows in the figure correspond to hierarchical clustering, while the bottom three rows correspond to K-means clustering.
Furthermore, we investigate the features input into the clustering process: the similarity between adjacent time periods (adj), the similarity to time period  0 (time0), and the upper triangular components (tri). 
From top to bottom, the rows represent the use of adjacent time periods, similarity to time period 0, and the upper triangular components in succession. 
We also examine the impact of standardization (z). Odd-numbered columns represent cases without standardization, while even-numbered columns represent cases with standardization.
The performance differences observed can be attributed to the presence or absence of standardization.

\subsection{t-SNE Visualization in Pseudo Data}
\label{sec:appendix:vis-pseudo}

\begin{figure*}[t]
    \centering
    \includegraphics[keepaspectratio, scale=0.07]{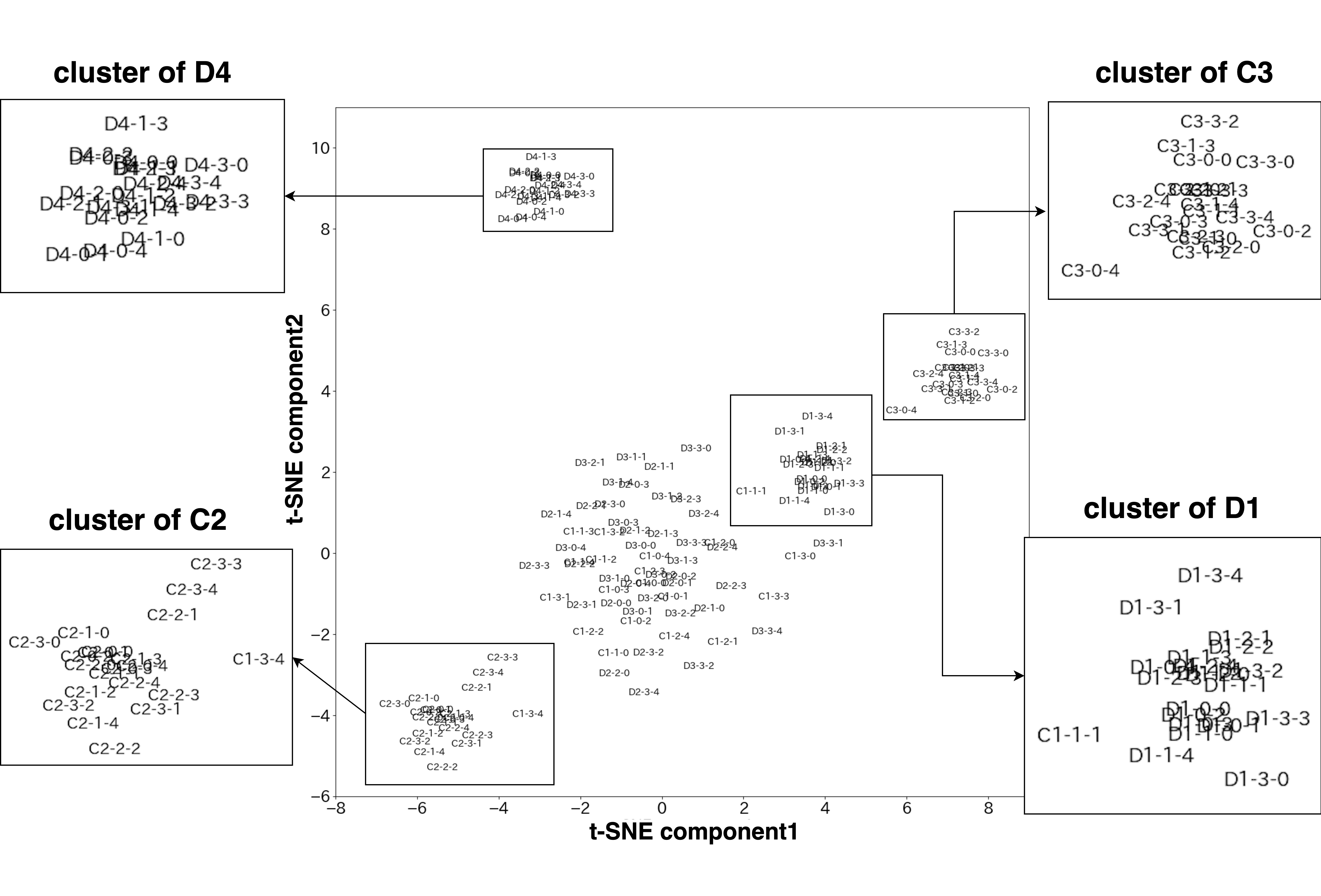}
    \caption{The result of visualizing the similarity matrices of pseudo-words in the pseudo-data using t-SNE in two dimensions. The input consists of the standardized upper triangular components of the similarity matrices, calculated using cosine similarity.}
    \label{fig:tsne-pseudo}
\end{figure*}

To examine how pseudo-words are distributed, we analyzed their visualization in two dimensions using t-SNE.
\autoref{fig:tsne-pseudo} shows the result of visualizing the similarity matrices of pseudo-words in two dimensions using t-SNE.
Some shifts patterns (C2, C3, D1, D4) are clearly separated into distinct clusters on t-SNE.
Compared to clustering on real data, the clusters in t-SNE are more distinctly separated, likely because the number of target words is smaller, with 20 pseudo-words prepared for each shifts pattern.

\section{Other Limitation}
One application in natural language processing is additional training for words that have undergone semantic shifts.
Pretrained large language models often rely on training data that becomes outdated over time, leading to a decline in performance when handling inputs reflecting the latest knowledge~\cite{lazaridou-2021-mind}. 
By tracking semantic shifts words, it becomes possible to identify and prioritize additional training for words that have shifted in meaning, mitigating performance degradation caused by the passage of time~\cite{su-etal-2022-improving,ishihara-etal-2022-semantic}.
Our method could be used to efficiently fine-tune pre-trained language models by prioritizing fine-tuning based on the identified patterns of semantic shifts. Additionally, varying the importance of words according to their semantic shifts patterns could further improve model adaptability. 
While our study focuses on the identification and analysis of patterns of semantic shifts, these promising applications highlight the broader potential of our approach and will be explored as part of future work.

\end{document}